\documentclass{article} % For LaTeX2e
\usepackage[preprint]{neurips_2025}
\pdfoutput=1

\usepackage{amsmath,amsfonts,bm}

\def\eqref#1{equation~\ref{#1}}
\def\1{\bm{1}}

\DeclareMathAlphabet{\mathsfit}{\encodingdefault}{\sfdefault}{m}{sl}
\SetMathAlphabet{\mathsfit}{bold}{\encodingdefault}{\sfdefault}{bx}{n}

\usepackage{microtype}
\usepackage{graphicx}
\usepackage{subcaption}
\usepackage{booktabs} % for professional tables
\usepackage{bbm}

\usepackage{comment}

\usepackage{etoc}
\etocdepthtag.toc{mtchapter}
\etocsettagdepth{mtchapter}{subsection}
\etocsettagdepth{mtappendix}{none}

\usepackage{hyperref}
\usepackage{url}
\usepackage{tikz}
\usetikzlibrary{trees}
\usepackage{multirow}
\usepackage{amsthm}
\usepackage{thmtools}
\usepackage{amsmath}
\usepackage{amssymb}
\usepackage{mathtools}
\usepackage{wrapfig} % Re-added for specific figure in original, but check usage
\usepackage[T1]{fontenc}
\usepackage{algorithm}
\usepackage{algpseudocode}

\usepackage{amsmath}
\usepackage{tikz}
\usetikzlibrary{decorations.pathreplacing,positioning}

\theoremstyle{plain}
\newtheorem{theorem}{Theorem}[section]

\newtheorem{lemma}[theorem]{Lemma}
\newtheorem{corollary}[theorem]{Corollary}
\theoremstyle{definition}
\newtheorem{definition}[theorem]{Definition}
\newtheorem{assumption}[theorem]{Assumption}
\theoremstyle{remark}

\definecolor{cadmiumorange}{rgb}{0.8, 0.33, 0.0}
\newcommand{\stress}[1]{\textcolor{cadmiumorange}{{#1}}}
\title{
When More is Less: \\
Understanding Chain-of-Thought Length in LLMs
}

\author{
Yuyang Wu\thanks{Equal Contribution} \\
Peking University\\
\And
Yifei Wang\textsuperscript{*} \\
MIT\\
\And
Ziyu Ye \\
University of Chicago \\
\And
Tianqi Du \\
Peking University\\
\And
Stefanie Jegelka \\
TUM\thanks{School of CIT, MCML, MDSI}~~and MIT\thanks{EECS and CSAIL} \\
\And
Yisen Wang\thanks{Corresponding Author: Yisen Wang (yisen.wang@pku.edu.cn)} \\
Peking University
}
\begin{document}
\maketitle

\begin{abstract}
Large Language Models (LLMs) employ Chain-of-Thought (CoT) reasoning to deconstruct complex problems. While longer CoTs are often presumed superior, this paper challenges that notion, arguing that \textbf{longer is not always better}. Drawing on combined evidence from real-world observations, controlled experiments, and theoretical analysis, we demonstrate that task accuracy typically follows an inverted U-shaped curve with CoT length, where performance initially improves but eventually decreases as the number of CoT steps increases. With controlled experiments, we further uncover the \textbf{scaling behaviors of the optimal CoT length}: it increases with task difficulty but decreases with model capability, exposing an inherent \textbf{simplicity bias} where more capable models favor shorter, more efficient CoT reasoning. This bias is also evident in Reinforcement Learning (RL) training, where models gravitate towards shorter CoTs as their accuracy improves. To have a deep understanding of these dynamics, we establish a simple theoretical model that formally proves these phenomena, including the optimal length's scaling laws and the emergence of simplicity bias during RL. Guided by this framework, we demonstrate significant practical benefits from training with optimally-lengthed CoTs and employing length-aware filtering at inference. These findings offer both a principled understanding of the "overthinking" phenomenon and multiple practical guidelines for CoT calibration, enabling LLMs to achieve optimal reasoning performance with adaptive CoTs tailored to task complexity and model capability.
% Chain-of-thought (CoT) reasoning enhances the multi-step reasoning capabilities of large language models (LLMs) by breaking complex tasks into smaller, manageable sub-tasks. Researchers have been exploring ways to guide models to generate more complex CoT processes to improve the reasoning ability of LLMs, such as long CoT and the test-time scaling law. However, for most models and tasks, does an increase in CoT length consistently lead to improved reasoning accuracy?
% In this paper, we observe a nuanced relationship: as the number of reasoning steps increases, performance initially improves but eventually decreases. To understand this phenomenon, we provide evidence that \textit{longer reasoning processes are increasingly susceptible to noise.} We theoretically prove the existence of an optimal CoT length and derive a scaling law for this optimal length based on model capability and task difficulty. Inspired by our theory and empirical findings, we conduct experiments on both synthetic and real-world datasets. Furthermore, we propose Length-filtered Vote to alleviate the effects of excessively long or short CoTs and demonstrate that training models with optimal-length CoT data can significantly improve performance. Our findings highlight the critical need to calibrate CoT length to align with model capabilities and task demands, offering a principled framework for optimizing multi-step reasoning in LLMs.
\end{abstract}

% \begin{quote}
% \end{quote}

\section{Introduction}
\begin{center}
\textit{``Everything should be made as simple as possible, but not simpler.''}~~ — Albert Einstein
% \vspace{-0.2in}
% \[
% \tikz[baseline=(A.base)]{
%   % First phrase + comma
%   \node[inner sep=0pt, outer sep=0pt] (A)
%     {\textit{Everything should be made as simple as possible},};
%   \draw[thick,decorate,decoration={brace,mirror,raise=2pt,amplitude=4pt}]
%     (A.south west) -- (A.south east)
%     node[midway,below=2pt,font=\footnotesize]
%       {optimal CoT length has a simplicity bias.\quad\quad\quad};

%   % Second phrase, positioned flush against the comma
%   \node[anchor=west,inner sep=0pt, outer sep=0pt, xshift=0.5em] (B)
%     at (A.east)
%     {\textit{but not simpler}.};
%   \draw[thick,decorate,decoration={brace,mirror,raise=2pt,amplitude=4pt}]
%     (B.south west) -- (B.south east)
%     node[midway,below=2pt,font=\footnotesize]
%       {\quad\quad \quad\quad \quad\quad optimal CoT length grows with task complexity.};

%   % Attribution, immediately after the period
%   \node[anchor=west,inner sep=0pt, outer sep=0pt, xshift=0.5em]
%     at (B.east)
%     {— Albert Einstein};
% }
% \]

\end{center}

% \section{Introduction}

Large language models (LLMs) have demonstrated impressive capabilities in solving complex reasoning tasks \citep{icl:Brown,touvron2023llama}. A key technique for its success is Chain-of-Thought (CoT) reasoning \citep{cot:Wei}. By generating explicit intermediate reasoning steps, CoT allows models to break down complex problems into simpler, more manageable sub-problems, akin to a divide-and-conquer strategy \citep{Divide-and-Conquer}. 
% This approach has led to significant advancements in multi-step reasoning. 

A common intuition, supported by some research \citep{complex-prompt,long-step}, is that longer and more detailed CoT processes generally lead to better performance, especially for difficult tasks. Meanwhile, recent observations also suggest that concise CoTs can sometimes be effective, albeit with potential performance trade-offs on complex problems \citep{ConciseCoT}. This raises a crucial question: does reasoning performance consistently improve as CoTs grow longer and longer? 

\begin{figure}[t]
\centering
\begin{subfigure}{0.49\textwidth}
    \centering
    \includegraphics[width=\linewidth]{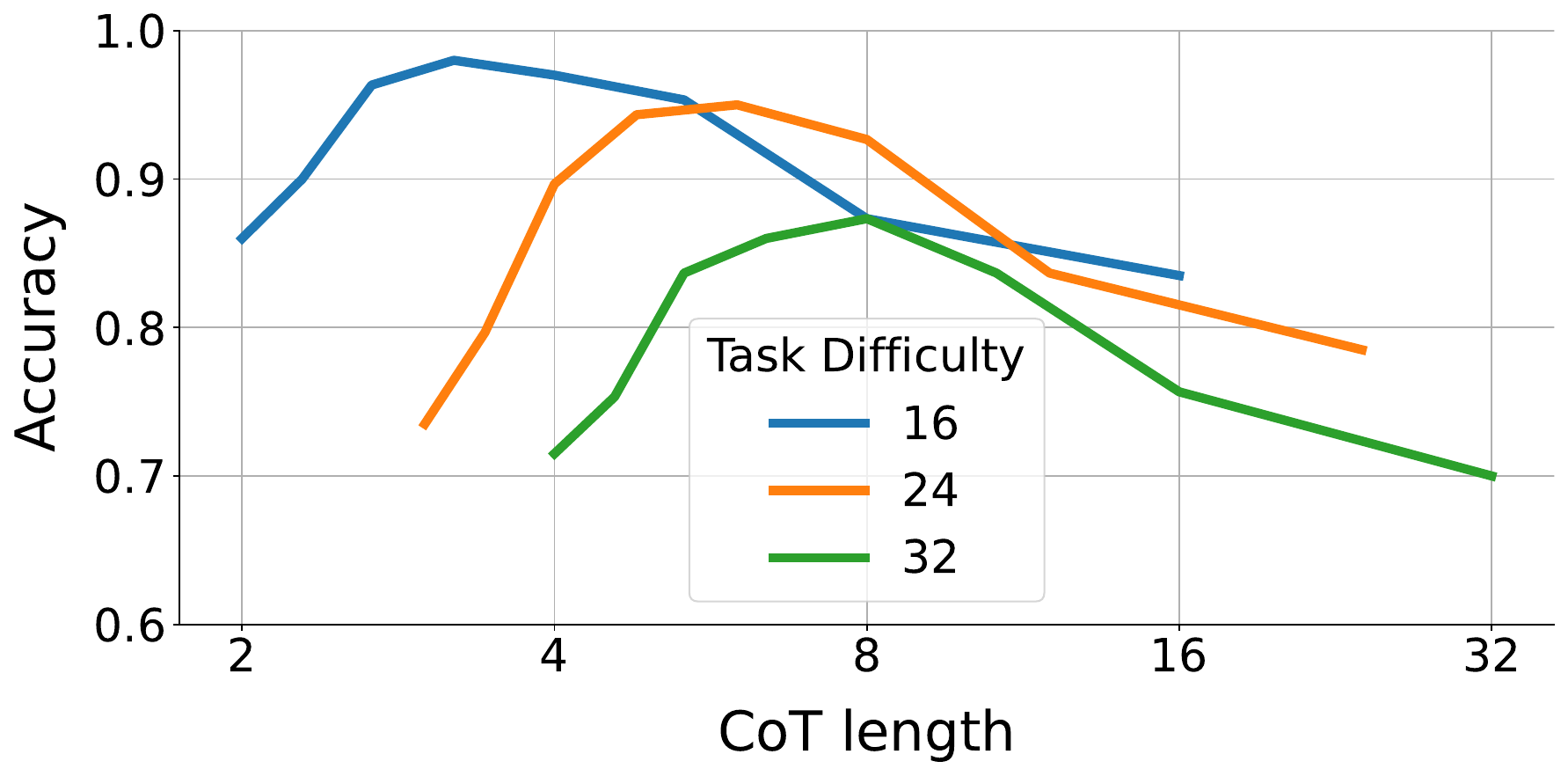}
    \caption{Reasoning Accuracy vs. CoT length}
    \label{fig:task-difficulty}
\end{subfigure}\hfill
\begin{subfigure}{0.49\textwidth}
    \centering
    \includegraphics[width=\linewidth]{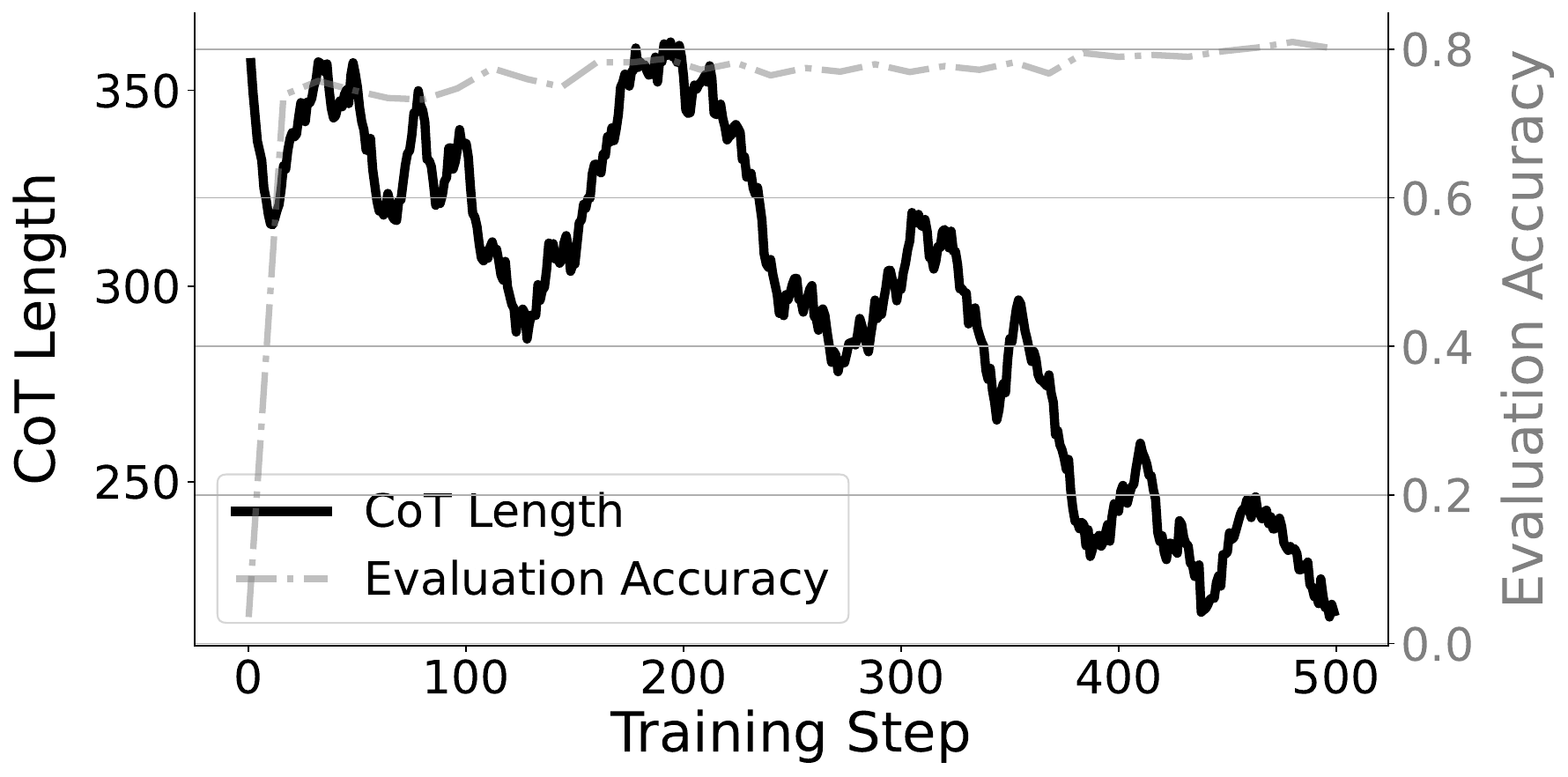}
    \caption{Evolution of LLMs' CoT lengths during RL}  \label{fig:real_world_rl_cot_decrease_conceptual}
\end{subfigure}
\hfill
\caption{
(a) The performance of a 6-layer GPT2 model (Section~\ref{sec:controlled_study}) follows inverted U-shaped curves on arithmetic tasks at different difficulty levels. As task difficulty increases, the accuracy peak progressively shifts toward longer CoT lengths.
(b) As RL training progresses and model accuracy on reasoning tasks improves, the average length of the generated Chain-of-Thought can decrease. This hints at the model learning more efficient, concise reasoning paths (\textit{i.e.}, simplicity bias). We conduct this experiment using Qwen2.5-7B-Instruct trained with GRPO on the LeetCode-2K dataset.}
\label{fig:Real_world_rl_experiments} % Overall figure label
\end{figure}

In this paper, through a comprehensive combination of evidence from theoretical analysis, controlled synthetic experiments, and real-world observations, we show that for CoT length, \textbf{longer is not always better}. As illustrated by the trend in Figure~\ref{fig:task-difficulty} , when plotting task accuracy against measures related to the CoT length, performance typically follows an \textbf{inverted U-shaped} curve. Performance initially improves as the CoT appropriately decomposes the task, but eventually deteriorates if the CoT becomes excessively long (increasing error accumulation) or too short (steps are too complex). This indicates the existence of an \textbf{optimal CoT length} that balances these competing factors.
% become 
% there exists an optimal  also validate 
% Performance initially improves as the CoT appropriately decomposes the task, but eventually deteriorates if the CoT becomes 
% excessively long 
% (increasing error accumulation) or too short (steps are too complex). 
% This indicates the existence of an \textbf{optimal CoT length} that balances these competing factors.

Further, we discover scaling behaviors of this optimal CoT length with respect to model capability and task difficulty: harder tasks tend to have longer optimal CoTs, while more capable models often achieve peak performance with shorter optimal CoTs. This latter point interestingly implies an inherent \textbf{simplicity bias} in LLM reasoning, where models favor shorter, more efficient reasoning paths as their capabilities increase. Moreover, we observe this simplicity bias during LLMs' reinforcement learning (RL) training. As shown in Figure
~\ref{fig:real_world_rl_cot_decrease_conceptual}, RL-trained models exhibit a gradual shift towards using shorter CoTs compared to the base model, 
% the overall CoT length can decrease towards the end of RL training as accuracy improves, 
indicating an acquired preference for shorter CoTs as a result of the simplicity bias of optimal CoT length. 
% Consequently, as seen in synthetic experiments (Figure~\ref{fig:rl_frequency_comparison}, i.e., Figure~\ref{fig:U-curve}b), RL-trained models may exhibit a gradual shift towards using shorter CoTs compared to the base model. 
This surprising phenomenon parallels humans' natural preference for simplest possible reasoning processes, as evident in Einstein's quote.

To gain a deeper understanding of the rise of optimal CoT length and its simplicity bias, we focus on a controlled study using a synthetic arithmetic task that allows us to ablate nuanced factors present in practical LLM training. In this controlled setting, we not only successfully replicate these phenomena but also theoretically derive the existence of the optimal CoT length and its scaling behaviors with respect to task complexity and model capability. Intuitively, task decomposition into more steps yields easier subtask but also accumulate errors exponentially, leading to an optimal tradeoff at an intermediate CoT length.
Notably, this theory also explains the emergence of the simplicity bias as observed during RL training. Thus, although simple, our theory provides valuable characterization of LLMs' behaviors during CoT. Translating this understanding into practice, we show significant benefits from training with optimally-lengthed CoTs and employing \emph{Length-aware Vote} to filter out excessively long CoTs at inference.

To summarize, this paper makes the following main contributions:
\begin{itemize}
    \item We demonstrate the existence of an optimal CoT length and the simplicity bias of CoT on both real-world LLMs (Section~\ref{sec:real_world_phenomenon}) and synthetic arithmetic experiments (Section~\ref{sec:controlled_study}).
    \item We establish a theoretical model of CoT that allows to formally characterize and prove the existence of an optimal CoT length as well as its scaling laws and simplicity bias (Section~\ref{sec:theoretical_analysis}).
    \item We explore the implications of these findings, showing how training with optimal-length CoT data can significantly boost performance, and how filtering excessively long CoTs with entropy measures can benefit reasoning performance at inference (Section~\ref{sec:implications_applications}).
    % we propose "Length-filtered Vote," an inference-time strategy to capitalize on optimal CoT length (Section~\ref{sec:implications_applications}).
\end{itemize}

Our findings offer a fresh perspective for calibrating CoT generation, moving beyond the assumption that longer is always better. By understanding and adapting to the optimal CoT length, we can develop LLMs that reason more effectively, avoiding both underthinking and counterproductive overthinking.
% The remainder of this paper is structured as follows: Section~\ref{sec:real_world_phenomenon} presents our real-world observations. Section~\ref{sec:controlled_study} details the controlled synthetic experiments. Section~\ref{sec:theoretical_analysis} outlines our theoretical model. Section~\ref{sec:implications_applications} discusses the implications and applications. Section~\ref{sec:related_work} reviews related literature, followed by the conclusion in Section~\ref{sec:conclusion}.

\section{Optimal CoT Length and Simplicity Bias in Real-World LLMs}
\label{sec:real_world_phenomenon}

To ground our investigation in practical scenarios, we first explore the relationship between CoT length and reasoning performance using publicly available LLMs. 
% These experiments aim to establish whether the hypothesized optimal CoT length and Simplicity Bias manifest in complex tasks.

\subsection{Scaling Behaviors of Optimal CoT Length in Real-World LLMs}
\begin{figure*}[t]
\centering

\begin{subfigure}{0.32\textwidth}
    \centering
    \includegraphics[width=\linewidth]{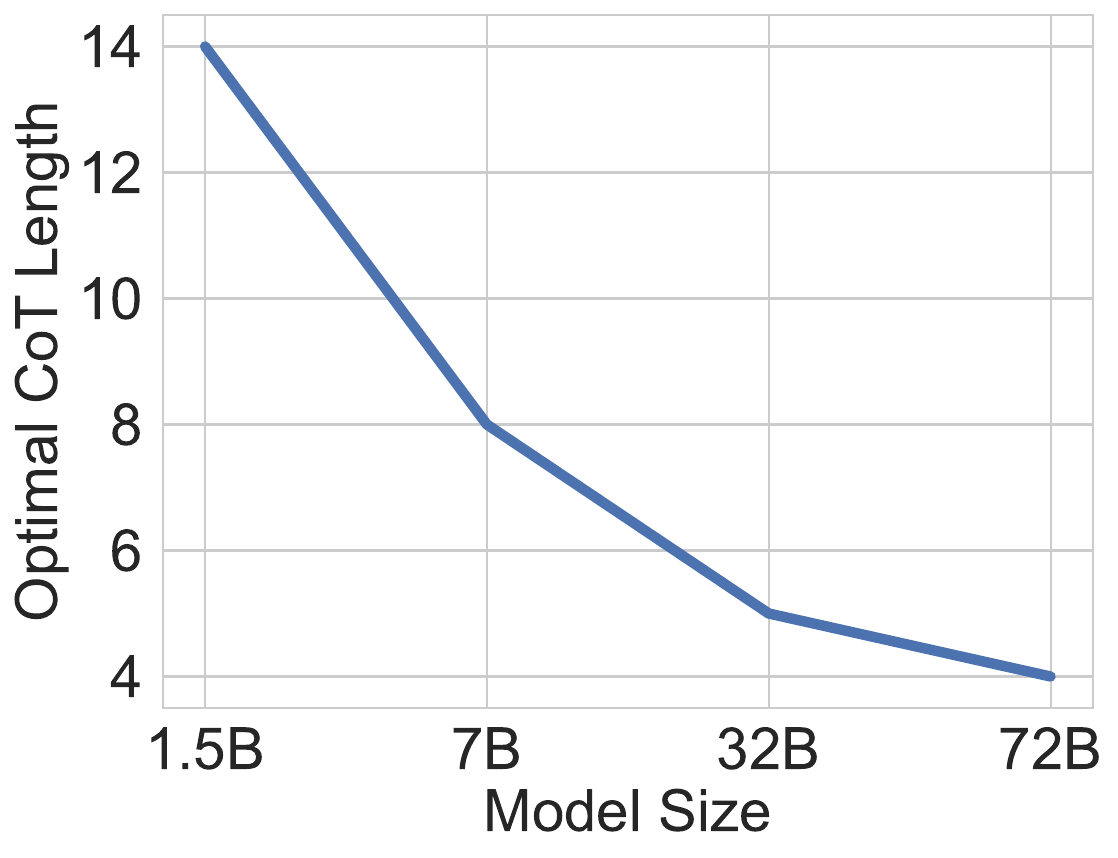}
    \caption{Optimal CoT length vs. Model size (Qwen2.5 series).}
    \label{fig:real_world_model_size_vs_opt_length}
\end{subfigure}%
\hfill
\begin{subfigure}{0.32\textwidth}
    \centering
    \includegraphics[width=\linewidth]{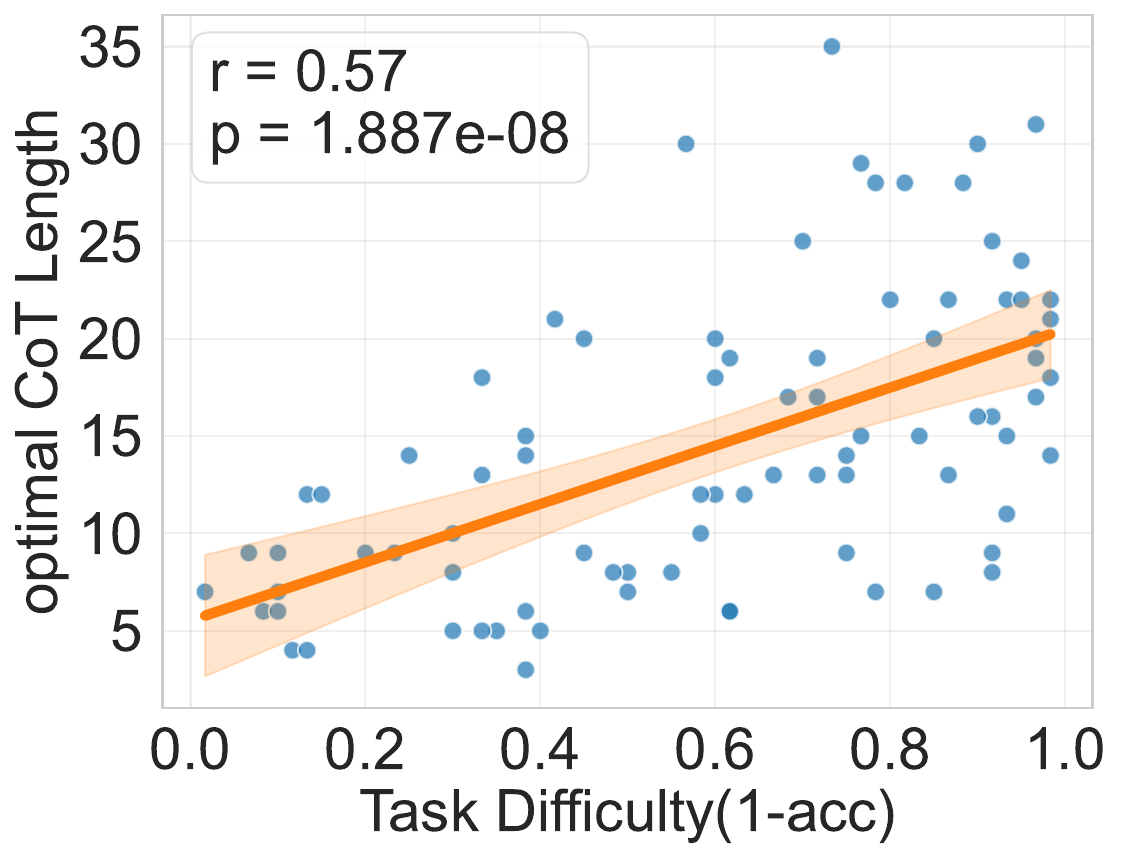}
    \caption{Optimal CoT length vs. Task difficulty (with the 1.5B model).}
    \label{fig:real_world_task_difficulty_vs_opt_length}
\end{subfigure}\hfill
\begin{subfigure}{0.32\textwidth}
    \centering
    \includegraphics[width=\linewidth]{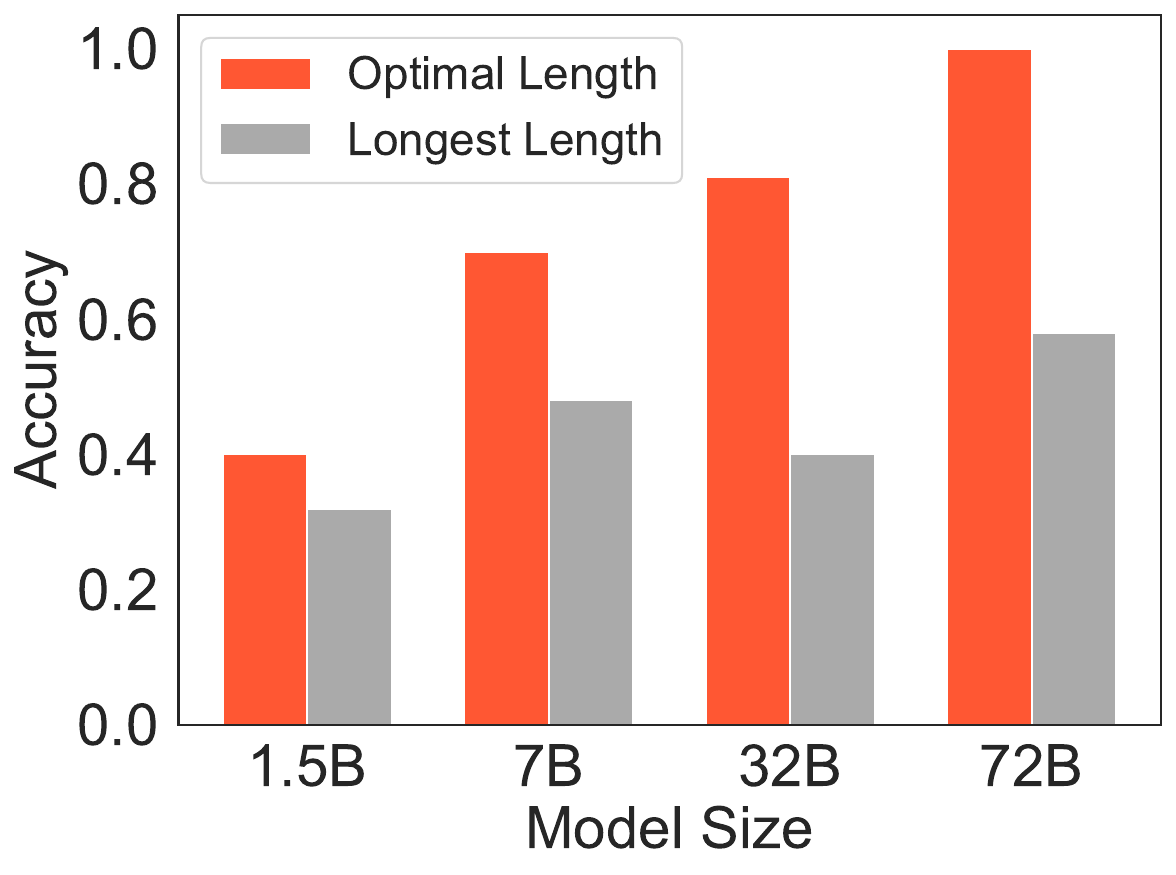}
    \caption{Optimal vs. Longest CoT length accuracy on MATH Level 5.}
    \label{fig:real_world_acc_gap}
\end{subfigure}%
\caption{Real-world CoT length observations. (a) Larger models tend to achieve optimal performance with shorter CoTs. (b) More difficult tasks (as measured by lower accuracy on the x-axis) tend to require longer optimal CoTs (with a positive correlation of significance $p \ll 0.05$). (c) Accuracy for CoTs of optimal length is significantly higher than that of the longest CoTs.}
\label{fig:real_world_observations}
\end{figure*}

\textbf{Setup.}
To assess how model capability interacts with CoT length. We evaluate Qwen2.5 series of Instruct models \citep{qwen} on Level 5 questions in MATH dataset composed of challenging competition mathematics problems~\citep{math}. For each question, we generate 60 solutions with as much variation in length as possible. The CoT length is determined by the number of intermediate reasoning steps generated by the model. The optimal CoT length is the one that yields the highest average accuracy. See Appendix~\ref{app:real_world} for additional experiments (MMLU STEM dataset~\citep{mmlu}, different models) and implementation details on step segmentation and solution length control.

\textbf{Optimal Length Decreases with Stronger Model Capabilities:} For each model, we randomly select 30 questions since our focus lies in exploring different lengths of solutions for the same problem rather than evaluating the whole dataset. As depicted in Figure~\ref{fig:real_world_model_size_vs_opt_length}, there is a clear trend where the optimal CoT length decreases as the model size increases. For instance, the optimal length shifts from 14 steps for the 1.5B parameter model to 4 steps for the 72B parameter model. This suggests that more capable models can consolidate reasoning into fewer, more potent steps, aligning with the Simplicity Bias concept where stronger models prefer shorter effective paths.

\textbf{Optimal Length Grows with Harder Tasks:} We also investigate how task difficulty influences the optimal CoT length. For this, we consider 100 randomly selected questions and compute the accuracy of an LLM on each question from $60$ sampled solutions. We use (1 - accuracy) on these questions as a proxy for the difficulty. Figure~\ref{fig:real_world_task_difficulty_vs_opt_length} shows a statistically significant positive correlation (notably $p = 1 \times 10^{-8} \ll 0.05$) between task difficulty and the optimal CoT length of Qwen1.5B-Instruct model. This indicates that more challenging problems will significantly benefit from a longer CoT with more extended decomposition steps. Similar trends for other models are provided in Appendix~\ref{app:difficulty}.

\textbf{Excessively Long CoTs Lead to Significant Degradation:} The above scaling behaviors of CoT suggest that one should adaptively select the optimal CoT length w.r.t.~the given model and task. Here, we illustrate the significance of this choice by compare the performance of using the optimal and the longest CoT lengths. As shown in Figure~\ref{fig:real_world_acc_gap}, there is a large gap between the two that grows larger as the models become more capable. For a 72B model, the gap can be as large as 40\% accuracy, showing great potential gains of adapting CoT length to  attain optimal reasoning performance.
% adaptively ch  selecting the CoT solutions with the \emph{longest} lengths does not consistently yield the best performance. Instead, peak accuracy is often found at an intermediate, or optimal, CoT length among all possible CoTs. This supports our initial hypothesis that excessively long CoTs can be detrimental.

\subsection{Simplicity Bias in Reinforcement Learning}
\label{sec:real_world_rl_observation}

A common belief in the ongoing development of advanced reasoning models is that reinforcement learning (RL) leads to more lengthy output in reasoning models. Nevertheless, recent studies \citep{qwen_vs_llama_rl} also revealed that RL-trained model behaviors remain largely depend on the base model. It is yet unclear what is the exact influence of RL training on CoT length.
% how CoT length
% closely tied to the base model, and observed length increases might be due to phenomena like extensive backtracking rather than more elaborate core reasoning.
% To have a 
To have a clear understanding of this process, we monitor the evolution of CoT length during GRPO training~\citep{grpo}, using LeetCode-2K~\citep{leetcodedataset} with Qwen2.5-7B-Instruct~\citep{qwen}. We refer readers to Appendix~\ref{app:exp-real} for additional details and ablations. 
% Our own observations in real-world scenarios 
As shown in Figure~\ref{fig:real_world_rl_cot_decrease_conceptual}, 
% reveal a fascinating counter-trend: 
% during RL fine-tuning aimed at improving reasoning accuracy (e.g., using outcome-based rewards), 
through optimizing outcome rewards from model rollouts, the average response length of RL models can decrease as training converges. As a result, RL-trained model has shorter CoTs (on average) than the base model, indicating that RL has a \emph{simplicity bias} that favors shorter answers instead of long answers. 
% the average length of the \emph{successful} or \textit{preferred} CoT paths generated by the model often \textit{decreases} over
% training iterations, even as overall task performance improves. This suggests that the RL process, by rewarding successful outcomes, may implicitly guide the model towards more concise and efficient reasoning strategies corresponding to the optimal CoT length. 
% This observation provides initial real-world evidence for an inherent simplicity bias, where systems learn to favor shorter, effective solutions. 
% We will revisit this phenomenon with controlled experiments in Section~\ref{sec:implications_applications}.

% \begin{figure}[t]
% \centering
% % \fbox{\rule{0pt}{2.5cm} \rule{0.6\linewidth}{0pt}}
% \includegraphics[width=0.6\linewidth]{Figures/Realworld/length_vs_score.pdf}
% \caption[(Corresponds to original Fig 1c)}
% \label{fig:real_world_rl_cot_decrease_conceptual}
% \end{figure}

\section{A Controlled Study of CoT Length in Arithmetic Tasks}
\label{sec:controlled_study}

The observations from real-world LLMs in Section~\ref{sec:real_world_phenomenon} suggest a complex interplay between CoT length, model capability, and task difficulty. However, real-world CoTs involve numerous uncontrolled variables (e.g., diverse reasoning strategies, planning, backtracking) and varying types of base model pre-training, making a precise mechanistic understanding challenging. To overcome these limitations and rigorously examine our hypotheses about optimal CoT length and Simplicity Bias, we develop a controlled experimental setup using synthetic arithmetic tasks.

\subsection{Experimental Setup}
\textbf{Dataset:} Our synthetic dataset consists of arithmetic problems involving only a sequence of addition operations. The inherent difficulty of a problem is quantified by the total number of addition operators, $T$. For any given problem with $T$ operators, we generate multiple valid CoT solutions, differing in their length and granularity. The CoT length $N$, is the number of intermediate reasoning steps. Each step $i$ in a CoT processes a certain number of operators, $t_i$. For simplicity in our controlled study, we structure solutions such that $t_i$ is (approximately) constant for all steps in a given CoT, denoted as the step size $t$ (operators per step), where $N \approx T/t$.

For example, consider an arithmetic problem like "$1+2+3+4+5+6+7$". This problem involves $T=6$ addition operators. We can construct different CoT solutions for this problem:
\vspace{-0.1in}
\begin{itemize}
    \item A \textit{long CoT solution} might be designed to process $t=1$ operator per step. This would result in $N=6$ reasoning steps.
% \vspace{-0.1in}    
    \begin{verbatim}
   Problem: 1+2+3+4+5+6+7
   Step 1: 1+2 = 3. (Remaining: 3+3+4+5+6+7)
   Step 2: 3+3 = 6. (Remaining: 6+4+5+6+7)
   ...
   Step 6: 21+7 = 28. (Final Answer)
    \end{verbatim}
\vspace{-0.2in}
    \item A \textit{shorter CoT solution} for the same problem might process $t=3$ operators per step. This would result in $N=2$ reasoning steps.
    \begin{verbatim}
   Problem: 1+2+3+4+5+6+7
   Step 1: 1+2+3+4 = 10. (Remaining: 10+5+6+7)
   Step 2: 10+5+6+7 = 28. (Final Answer)
    \end{verbatim}
\end{itemize}
\vspace{-0.2in}
This dataset design is crucial as it allows us to systematically vary the CoT length ($N$) or the number of operators processed per step ($t$) for problems of a fixed total difficulty ($T$). This enables a focused study on how the structure of the reasoning process itself impacts performance. More discussion of problem definition, data format, CoT generation, and considerations for choosing task data formatting and task design, is provided in Appendix~\ref{app:setup}.

\textbf{Model and Training:} We train GPT-2 models \citep{radford2019language} of varying depths (number of layers), keeping other hyperparameters fixed. Model depth is known to be a significant factor representing model capabilities for reasoning tasks \citep{cot-theory:allen-zhu,transformer_layer}. Controlling this hyperparameter alone allows us to study the impact of model capability on optimal CoT length.
% \textbf{Training and Testing:} 
Models are trained with CoT solutions that can be automatically synthesized for this task, with varying total operators $T$ and CoT lengths $N$ (or equivalently the step sizes $t$). For testing, we can guide the model to produce a CoT of a specific length (e.g., by prompting with a control token indicating the desired number of operators $t$ per step) or allow it to choose its preferred length. Further details are in Appendix~\ref{app:syn}.

\begin{figure}[t] % Use figure* for a page-wide figure, suitable for 3 subplots
\centering

\hfill % Horizontal space
\begin{subfigure}{0.32\textwidth}
    \centering
    \includegraphics[width=\linewidth]{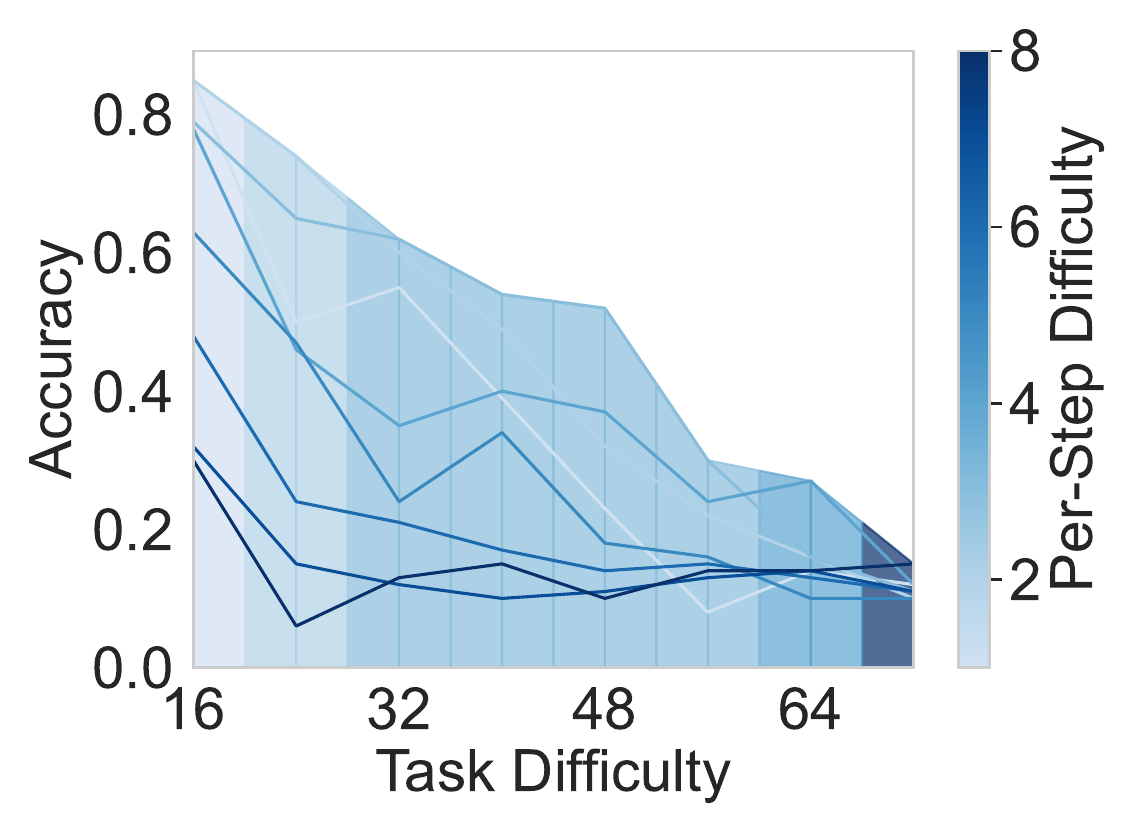}
    \caption{Optimal per-step difficulty}
    \label{fig:envelope_synthetic}
\end{subfigure}
\hfill % Horizontal space
\begin{subfigure}{0.32\textwidth}
    \centering
    \includegraphics[width=\textwidth]{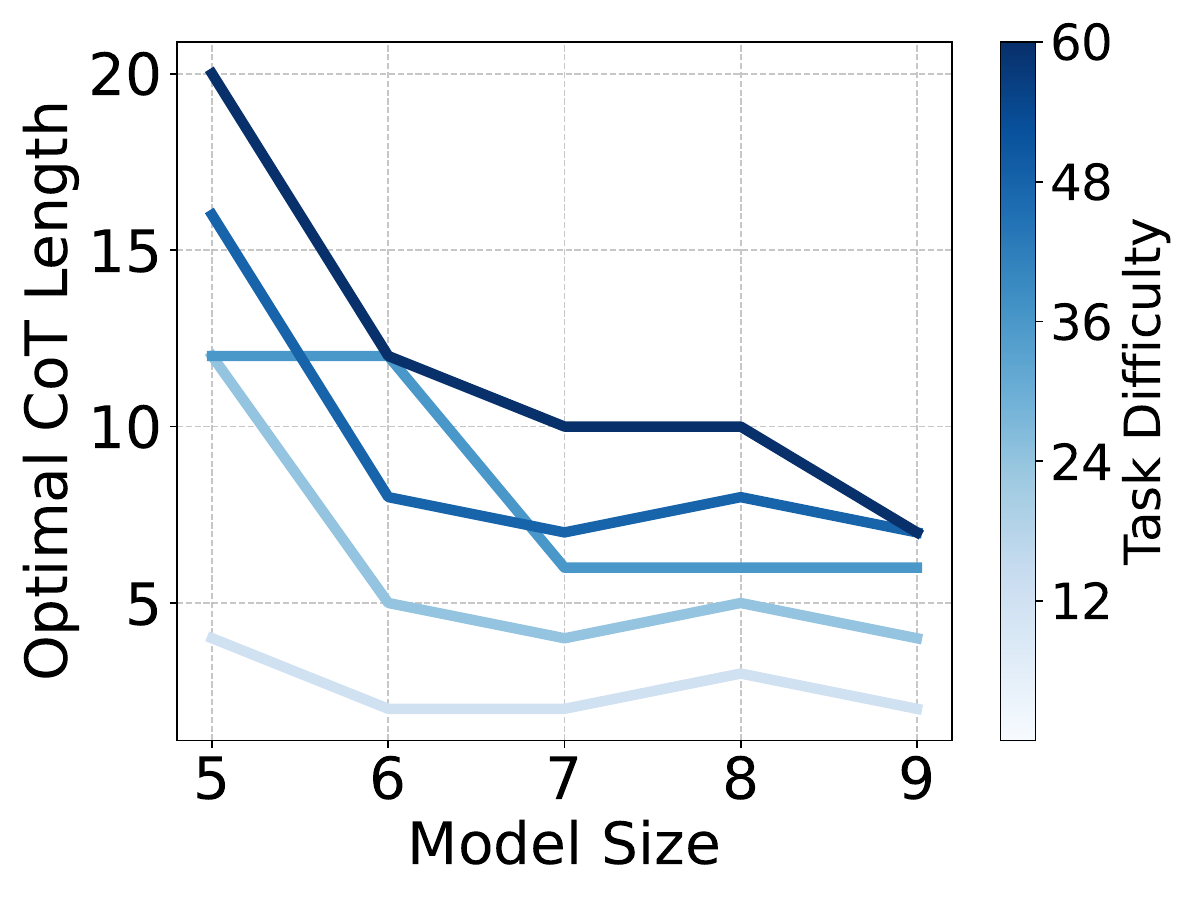}
    \caption{Optimal CoT vs. model size}
    \label{fig:model-size}
\end{subfigure}
\begin{subfigure}{0.32\textwidth} 
\vspace{-10pt}
    \centering
    \includegraphics[width=\linewidth]{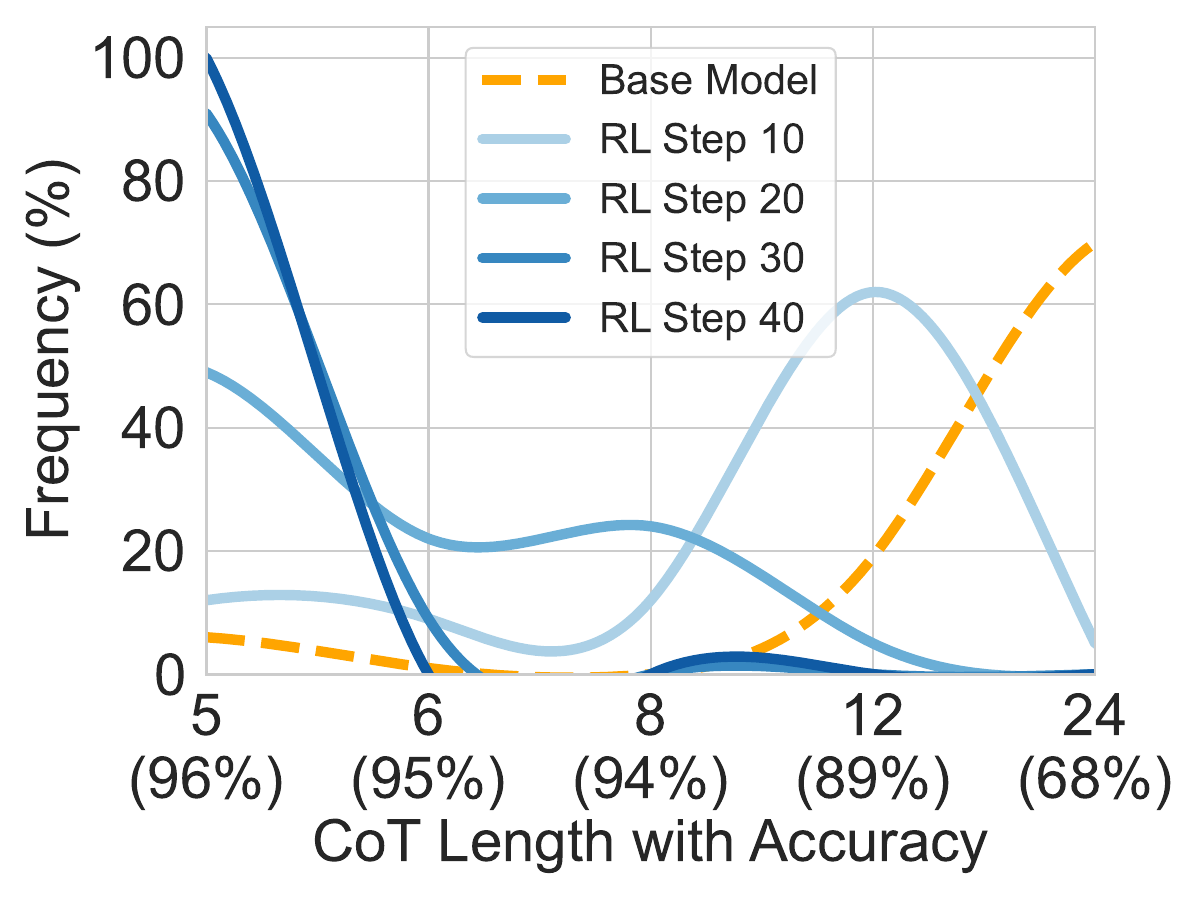}
    \caption{Evolution of CoT during RL}
    \label{fig:rl_freq_synthetic}
% \vspace{-10pt}
\end{subfigure}
\caption{
\textbf{CoT Behaviors in Synthetic Experiments:}
(a) Each curve corresponds to a specific CoT strategy with fixed per-step difficulty. The color of the bar beneath each curve represents the optimal per-step difficulty ($t$) at each task difficulty. 
% As task difficulty increases, the envelope formed by the upper bound of the curves reveals the adaptation of optimal reasoning strategies.
The progressively darker gradient colors indicates that harder tasks consistently favor higher per-step difficulty.
(b) Change of the optimal CoT length with increasing model size across different task difficulty levels. 
% Horizontally, a
As model size increases, the optimal CoT length decreases. For a fixed model size, harder tasks also exhibit longer optimal CoTs.
(c) During RL training, the model policy gradually favors a shorter CoT that corresponds to the optimal length.}
\label{fig:combined_synthetic_optimal_cot_analysis}
\vspace{-20pt}
\end{figure}

% \begin{wrapfigure}{r}{0.45\textwidth} 
% \vspace{-10pt}
%     \centering
%     \includegraphics[width=\linewidth]{Figures/Synthetic/rl_frequency_comparison_curve_long.pdf}
%     \caption{RL frequency shifting: Model preference shifts towards optimal step complexity ($t^*$) during PPO training for a fixed task/model (here, converging to $t=5$).}
%     \label{fig:rl_freq_synthetic}
% \vspace{-10pt}
% \end{wrapfigure}

\subsection{Scaling Laws of the Optimal CoT Length and and Practical Insights}
\label{sec:practical_insights_synthetic} % Added a label for the subsection

Our controlled experiments not only corroborate the CoT behaviors observed in real-world scenarios but also allow for a more fine-grained analysis. These findings uncover several key scaling behaviors of the optimal CoT length that shed light into the practical designs of LLM reasoning.
% for designing and training LLMs for reasoning.

\textbf{I. Harder-Tasks' CoTs Peak at Longer Lengths (\stress{Adaptive CoT Length Matters}):}
Our synthetic experiments further confirm the existence of an optimal CoT length, which manifests itself as an inverted U-shaped performance curve when plotting accuracy against the number of reasoning steps, as shown in Figure~\ref{fig:task-difficulty}. This clearly indicates that both "underthinking" (CoT too short) and "overthinking" (CoT too long) are detrimental, underscoring the critical benefit of generating CoTs with adaptive lengths tailored to the problem's demands. Moreover, we observe that the optimal CoT length shifts right as the task difficulty $T$ gets larger, indicating that solving a harder task optimally requires a longer CoT (also observable numerically from Figure~\ref{fig:model-size}). This suggests that a good reasoning model should be able to vary CoT lengths w.r.t.~the overall task complexity.
% (from the introductory Figure~\ref{fig:U-curve}a)\yw{check}, for tasks with fixed difficulty $T \in \{16, 24, 32\}$ and a given model capability (e.g., 6 layers), performance varies significantly with CoT structure. When the number of operators processed per step ($t$) is very small (leading to a very long CoT with many steps, $N$), accuracy is low due to error accumulation. Conversely, when $t$ is very large (leading to a very short CoT with few steps, $N$), accuracy also drops because each step becomes too complex for the model to solve reliably. Peak performance is achieved at an optimal intermediate $t$ (and thus optimal $N$). 

\textbf{II. Harder Tasks Peak at Harder Sub-tasks (\stress{Adaptive Per-Step Computation Helps}):}
Figure~\ref{fig:envelope_synthetic} %(part of Figure~\ref{fig:combined_synthetic_optimal_cot_analysis}a) 
illustrates how the number of operators per step ($t$) impacts model accuracy across varying task difficulties ($T$). The envelope curve, tracing peak performance, reveals that as tasks become more challenging (larger $T$), optimal performance is often achieved by CoTs that involve more complex computations \emph{per step} (i.e., a larger optimal $t^*$). This suggests that for harder problems, simply increasing the number of simple steps may not be as effective as increasing the complexity of each sub-task the model tackles within the CoT. Current LLMs with fixed Transformer layers have limited intrinsic ability to adapt their per-step computational depth for different sub-tasks. This implies that their reasoning strategy might remain suboptimal. In contrast, recent advancements like looped Transformers, which enable adaptive recurrent depth \citep{geiping2025scaling,inner_thinking_transformer}, could offer a more promising avenue for dynamically adjusting per-step computation to align with this observed need, potentially leading to better reasoning performance.

\textbf{III. Stronger Models Achieve Optimal Performance with Shorter CoTs (\stress{Model-Aware CoT Data Matter}):}
We also examine how model capability (number of layers) influences the optimal CoT length. Figure~\ref{fig:model-size} %~\ref{fig:heatmap_synthetic} (part of Figure~\ref{fig:combined_synthetic_optimal_cot_analysis}b), 
indicates that, across different task complexities, the optimal number of CoT steps ($N^*$) consistently decreases as the model's capability (number of layers) increases. This is because stronger models can effectively handle more complex sub-tasks in each step, thus requiring fewer overall steps to reach the solution optimally. This finding has significant implications for training data curation. It suggests that to achieve peak performance, models of different sizes or capabilities require CoT data tailored to their respective optimal per-step complexities. Current practices, such as using the same CoT datasets to train LLMs of varying sizes or directly distilling CoTs from large models to small ones without adapting complexity, may be suboptimal. For instance, a small model might struggle to learn effectively from overly complex CoT demonstrations designed for a larger model. Our analysis advocates for training each model with CoT data of adaptive complexity, aligned with its specific capabilities, to help it reach its optimal reasoning performance.

\textbf{IV. RL Training Converges to Optimal CoT Length (\stress{RL Calibrates Reasoning Behaviors}):}
As discussed in Section~\ref{sec:real_world_rl_observation}, RL training of LLMs leads to shorter CoT lengths. Our synthetic experiments further replicate this phenomenon. We take a GPT-2 model pre-trained on CoT solutions of equally mixed lengths for a task of difficulty $T=24$ and apply RL using rule-based outcome rewards with PPO on VERL \citep{ppo,verl}. Figure~\ref{fig:rl_freq_synthetic} %(part of Figure~\ref{fig:combined_synthetic_optimal_cot_analysis}c) 
shows the change of the sampled CoT lengths along RL: as training progresses, the model increasingly favors the CoT structure corresponding to the optimal length $N^*=5$ that yields the peak accuracy ($96\%$) on this task.
% \sj{looks more like it favors $t=5$ or even 6?} 
This demonstrates that RL, by optimizing for task success, can implicitly guide the model's CoT generation policy towards the optimal length regime, thereby exhibiting the \emph{simplicity bias}.
This offers a fresh perspective for understanding the benefits of RL in LLM training: even if the initial CoT data used for pre-training or supervised fine-tuning is suboptimal (e.g., misaligned with the model size or the task complexity), RL can help calibrate the model's behavior towards generating more optimally-lengthy CoTs.
% The effectiveness of the RL calibration would depend on the coverage of the base model and the efficiency of RL training.

\section{Theoretical Analysis: Why an Optimal CoT Length Exists}
\label{sec:theoretical_analysis}

The empirical findings from both real-world and synthetic datasets consistently point to the existence of an optimal Chain-of-Thought (CoT) length. In this section, we provide a theoretical framework to explain this phenomenon, formalizing how factors like task decomposition and error accumulation interact to determine this optimal length, and how it scales with model capability and task difficulty. All proofs are deferred to Appendix~\ref{app:proof}.

\subsection{Theoretical Formulation}
\label{sec:theory}
Akin to the arithmetic tasks we studied in Section \ref{sec:controlled_study}, we use the following simple theory model to describe the CoT process.\footnote{Note that we do not explicitly model various cognitive reasoning behaviors (reflection, verification, backtracking) but instead regard them as one of the many ways that one can decompose a task into subtasks to ease problem solving, which can be understood as a part of our task decomposition formulation in a general sense.}
Let $N \in \mathbb{N^+}$ be the total number of steps in the CoT process. Let $T$ denote the total number of operators in the given arithmetic task (a proxy for task difficulty). We assume that each CoT step consists of a sub-question $q_i$ (e.g., $2+1=$) and its answer as $a_i$ (e.g., $3$). 

\begin{definition}[CoT Process Probability]
Given a task $q$ with $T$ total operators and a model $\theta$, the probability of an $N$-step CoT that leads to a final answer $a_{\text{final}}$ is:
% (all CoT included) can be factorized as:
% has the follo probability:
$$P(a_{\text{final}}|q,\theta,N) = \prod^N_{i=1}\underbrace{P(q_i|H_{i-1},q,\theta,N)}_{\text{sub-question}}\underbrace{P(a_i|q_i,H_{i-1},q,\theta,N)}_{\text{sub-answer}},$$
where $H_{k}:=[t_1,a_1,\cdots,t_k,a_k]$ denotes the CoT history of the first $k$ steps.
\end{definition}

Let % \( q_i \) the subtask given the history reasoning steps \( H_{i-1} \), total task \( q \), and total step number \( N \), and 
\( a_i^* \) denote the correct answer to subtask \( q_i \) and $q_i^*$ denote the unique correct sub-question for simplicity. To estimate the final accuracy \( A(N) = P(a_N = a_N^*|q, \theta) \), we need to estimate the sub-question accuracy \( P(q_i = q_i^*|H_{i-1}, q, \theta) \) and the sub-answer accuracy \( P(a_i = a^*|q_i, H_{i-1}, q, \theta) \).

For the \textbf{sub-question accuracy}, 
% as observed in our experiments, the loss for tokens appearing in \( t_i \) remains almost constant across different values of \( N \) and model sizes 
following experimental observation
(in Appendix~\ref{app:subtask}), we assume that the error rate of generating each question \( q_i \), denoted by \( \sigma(T) \in [0,1)\), is positively correlated with the total number of operators \( T \). Intuitively, as the number of operators increases, extracting the correct subtask becomes more challenging.  For the \textbf{sub-answer accuracy}, it is clear that when given subtask \( q_i \), \( P(a_i = a^*|q_i, H_{i-1}, q, \theta) \) is independent of the history reasoning steps \( H_{i-1} \) and is only influenced by the model \( \theta \) and the difficulty of the subtask \( q_i \). For each model, we define its capability \( M \) based on the reasoning boundary \citep{reasoningboundary}, e.g., the maximum number of operators the model can directly solve per step; thus, a stronger model has a larger $M$.
% Specifically, we consider \( M \) as the maximum number of operators the model can compute in a single step without CoT. 
% \begin{equation}
% \label{eq:M}
%     M = M(\theta)=\max_{t}\left\{P(a_i=a^*|t_i,\theta ) > \tau,  |t_i|=t\right\},
% \end{equation}
% where $|t_i|$ refers to the number of operators in subtask $t_i$ and $\tau$ denotes the minimum accuracy threshold which a model is considered to be capable to solve a problem. In the following discussion, we focus only on CoT chains that do not exceed the model's capability, which is $t < M$. 
We define the error rate of each subtask answer as \( E(N, M, T) \in [0,1]\).
% and thus we have 
% \begin{equation}
% \label{eq:task_ans_acc}
%     P(a_i = a^*|t_i, H_{i-1}, q, \theta) = 1 - E(N, M, T).
% \end{equation}
\begin{restatable}{proposition}{finalacc}
The total accuracy of $N$-step reasoning is 
\begin{equation}
    A(N) = P(a_{\text{final}} = a_{\text{final}}^*|q,\theta,N) = \alpha\left((1-E(N,M,T))(1-\sigma(T))\right)^N,
    \label{eq:total-acc}
\end{equation}
where \(\alpha\) denotes a constant value independent of \(N\).
\label{pro:general_acc}
\end{restatable}

Proposition~\ref{pro:general_acc} establishes the quantitative relationship between CoT's final performance $A$ and reasoning length \(N\). Once we obtain estimates for \(E(N,M,T)\) and \(\sigma(T)\), we can determine the optimal CoT length $N^*$ as a function of the model capability $M$ and task complexity $T$. 
% For simplicity, in the following discussion, we allow \( N \), \( M \), and \( T \) to be real numbers. Below, we begin by a simple case with a linear error rate to illustrate the basic idea, while a more generalized analysis with random variables and nonlinear relationship can be found in Appendix~\ref{}.
% a  In Section~\ref{sec:linear}, we analyze the case of a linear error rate, while in Section~\ref{sec:general}, we explore more general error functions.

\textbf{Simple Case with Linear Error.}
% \label{sec:linear_error_theory}
To gain intuition, we first consider a simple case where the sub-question error scales linearly with $T$, i.e., $\sigma(T) = T/C$, where $C$ is a constant representing the maximum task difficulty models are trained to handle. Throughout this analysis, we assume $\sigma(T) \leq 0.9$ to restrict our discussion to tasks that are within the model's training regime. Otherwise, it would be unreasonable to claim the model has learned to solve such problems.
% consider $T \in [0, 0.9C]$ so $1-\sigma(T) \in [0.1, 1]$.
We also assume the sub-answer error rate scales linearly with harder tasks, fewer steps, and weaker models as $E(N, M, T)= (T/N)/M = T/(NM)$. 
% This represents the ratio of operators per step to the model's single-step capability $M$. We require $1 - T/(NM) > 0$. 
Under these simplified conditions, we can derive the following closed-form expression for the optimal CoT length.
\begin{restatable}[Optimal CoT Length]{theorem}{lineartheory}
\label{thm:simple_optimal_N_theory}
    For a given model capability $M$ and task difficulty $T$, the total accuracy 
    % following op assumptions, the total accuracy becomes $A(N)$ becomes 
% \begin{align}
$A(N) = \alpha[(1-T/C )\cdot(1-T/(NM))]^N$ 
% \end{align}
 % as defined in 
 (Eq.~(\ref{eq:total-acc}))
 initially increases and then decreases as $N$ increases (forming an inverted U-shape). Thus, there exists an optimal CoT length:
    \begin{equation}
        N^*(M,T) = \frac{TZ}{M(Z+1)},
    \end{equation}
    that maximizes $A(N)$, where $Z = W_{-1}(-{1 - T/(Ce)} )$, and $W_{-1}(x)$ is the smaller real branch of the Lambert W function satisfying $w e^w = x$, and $e$ is the natural number.
\end{restatable}

This theorem formally establishes the inverted U-shaped curve and provides an explicit form for $N^*$. From this, we can formally prove the first three scaling behaviors characterized in Section~\ref{sec:practical_insights_synthetic}.

\begin{restatable}[Scaling laws of Optimal CoT Length]{corollary}{Scalinglawsforlinear} Based on Theorem~\ref{thm:simple_optimal_N_theory}, one can derive:
\begin{itemize}
    \item $N^*(M, T)$ increases monotonically with $T$, i.e., harder tasks require more reasoning steps to attain the optimal performance.
    \item The optimal number of operators per step $t^* = T/N^*(M,T) = M(1+1/Z)$ increases monotonically with $T$. This aligns with the envelope curve result (Figure~\ref{fig:envelope_synthetic}).
    \item  $N^*(M, T)$ decreases monotonically with $M$, i.e., more capable models require fewer reasoning steps to attain the optimal performance, reflecting the simplicity bias.
\end{itemize}
\label{coro:scaling_law_for_linear}
\end{restatable}

% \begin{corollary}[Optimal N vs. Task Difficulty]
% \label{cor:NwithT_theory}
% \end{corollary}

% \begin{corollary}[Optimal N vs. Model Capability]
% \end{corollary}
\textbf{Extension to Broader Scenarios.} Here, we adopt a simple linear model to facilitate intuitive understanding. However, this analysis can be extended to more general settings, including general error functions (\textit{only} with mild assumptions of monotonicity and convexity) and stochastic error models, where \textit{each subtask may exhibit a different error rate}. These extensions introduce additional technical subtleties but follow the same underlying principles. We defer this part to Appendix~\ref{app:extension}. 

\subsection{Why does RL Exhibit Simplicity Bias?}
\label{sec:rl_convergence}
% At last, let us take a look at 
The analysis above also provides a natural understanding of RL's simplicity bias (Section~\ref{sec:practical_insights_synthetic}).As in the arithmetic task, we generate samples within a finite discrete action space $\mathcal{A} = \{N_1, N_2, \dots, N_k\}$  during RL  that receive binary outcome rewards. 
% In the post-training stage, when applying an RLVR algorithm (e.g.\ PPO or GRPO) with only outcome supervision (reward $r=1$ for a correct final answer, $r=0$ otherwise), the choice of CoT length can be viewed as . 
This reduces to a stateless bandit: each
$N_i$ yields reward $r\in\{0,1\}$ with probability $A(N_i)$ (from
Proposition~\ref{pro:general_acc}). Let us parameterize a softmax policy
$\pi_\theta(N_i)=\frac{e^{\theta_i}}{\sum_{j}e^{\theta_j}},$
and define the RL objective as
$J(\theta)=\sum_{i=1}^k\pi_\theta(N_i)\,A(N_i).$
As a result, the policy-gradient becomes
$\nabla_{\theta_i}J=\sum_{j=1}^kA(N_j)\,\pi_\theta(N_j)\bigl(\delta_{ij}-\pi_\theta(N_i)\bigr).$

\begin{restatable}[RL Converges to Optimal CoT Length]{corollary}{rlconverge}
\label{cor:rl_converge}
For gradient ascent on $J(\theta)$ with sufficiently small step size, the policy
converges to a deterministic policy
$\pi_\theta(N_i)=1\quad\text{iff}\quad i=\arg\max_jA(N_j).$
Thus, RL training converges to the optimal CoT length $N^*=\arg\max_{N\in\mathcal{A}}A(N)$.
\end{restatable}
This corollary shows that RL will automatically discover the optimal length (usually shorter length) through optimizing the reward function and exhibit a decreasing CoT length as in the simplicity bias phenomenon. In this way, our theory offers an explanation of the optimal CoT length, its scaling behavior and RL's simplicity bias within a unified framework.

% In plain terms, each candidate chain‑of‑thought length can be thought of as a strategy that succeeds with its own probability. During RLVR post‑training, the algorithm repeatedly rewards successful lengths and down‑weights less successful ones. Even small differences in accuracy cause the policy to tilt more and more toward the best‑performing length. Over time, this reinforcement process concentrates the model’s choices on the one length that maximizes its chances of a correct answer.

% As a result, the LLM’s response length naturally converges to the optimal chain‑of‑thought length, because that is the option most consistently rewarded during training.  Exact gradient flow on the
% softmax parameters drives the policy toward the argmax action.

\section{Practical Applications of Optimal CoT Length}
\label{sec:implications_applications}

Guided by the understanding above, in this section, we illustrate via some proof-of-concept experiments that adapting LLM training and inference configurations to the optimal CoT length can improve the model's reasoning performance.
% we can benefit a lot from .
% Understanding the existence of an optimal CoT length and the associated Simplicity Bias has significant practical implications for training, fine-tuning, and deploying large language models for reasoning tasks.

\subsection{Training with Data of Optimal CoT Length}

% \begin{figure}[ht]
% \centering
% % \begin{subfigure}{0.48\textwidth}
%     \centering
%     \includegraphics[width=.48\linewidth]{Figures/Synthetic/rand_base_vs_opt.pdf}
%     \caption{Performance of models pre-trained on datasets with random CoT lengths (\textit{Base}) vs. optimal CoT lengths (\textit{Opt}) for their capability and task difficulty.}
%     \label{fig:optimal_training_synthetic}
% % \end{subfigure}
% % \caption{Learning and leveraging optimal CoT length. (a) Reinforcement learning naturally shifts model preference towards optimal-length CoTs. (b) Training models on curated data containing primarily optimal-length CoTs significantly improves performance.}
% % \label{fig:learning_optimal_cot}
% \end

\begin{figure*}[t] % Use figure* for a page-wide figure, suitable for 3 subplots
\centering
\begin{subfigure}{0.32\textwidth}
    \centering
    \includegraphics[width=\linewidth]{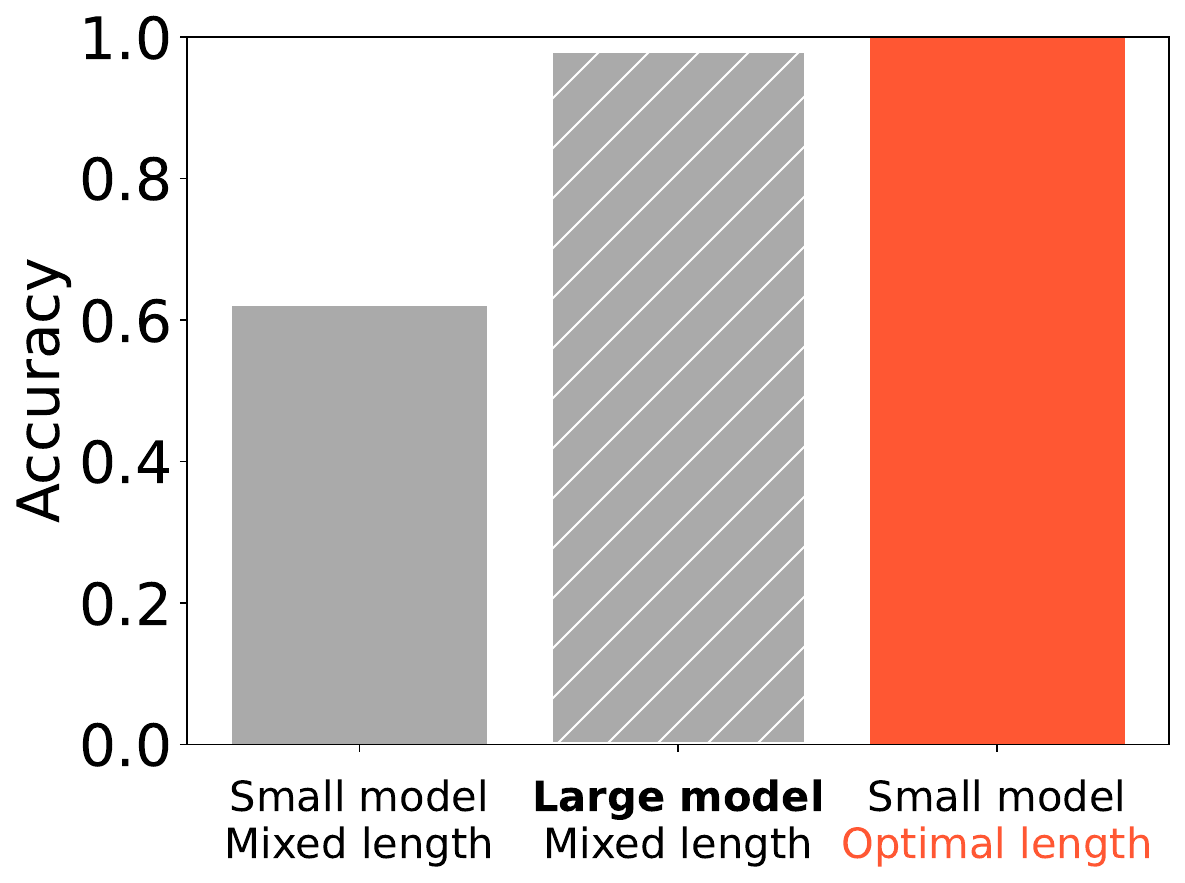}
    \caption{Influence of CoT Data ($T=32$)}
    \label{fig:opt_vs_rand_32}
\end{subfigure}
\hfill % Horizontal space
\begin{subfigure}{0.32\textwidth}
    \centering
    \includegraphics[width=\textwidth]{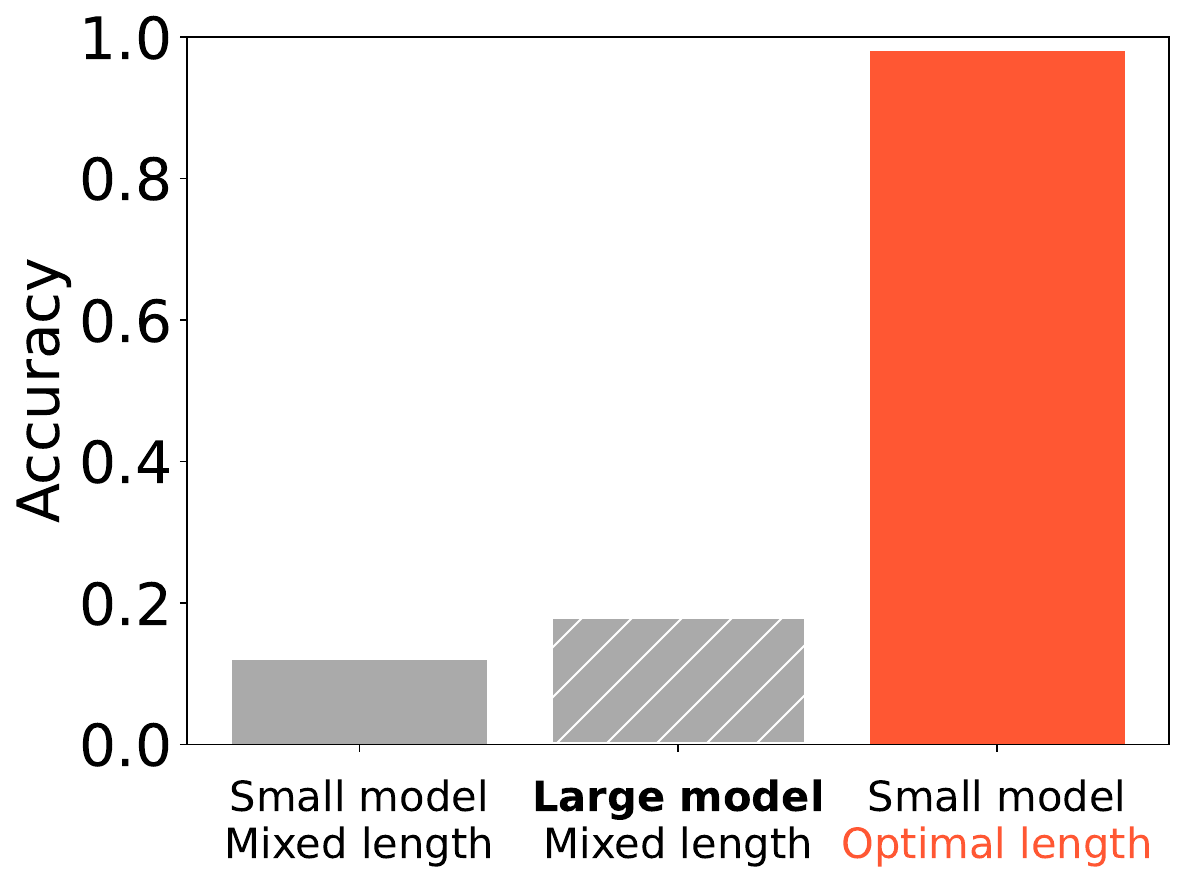}
    \caption{Influence of CoT Data ($T=64$)}
    \label{fig:opt_vs_rand_64}
\end{subfigure}
\hfill % Horizontal space
\begin{subfigure}{0.32\textwidth}
    \centering
    \includegraphics[width=\linewidth]{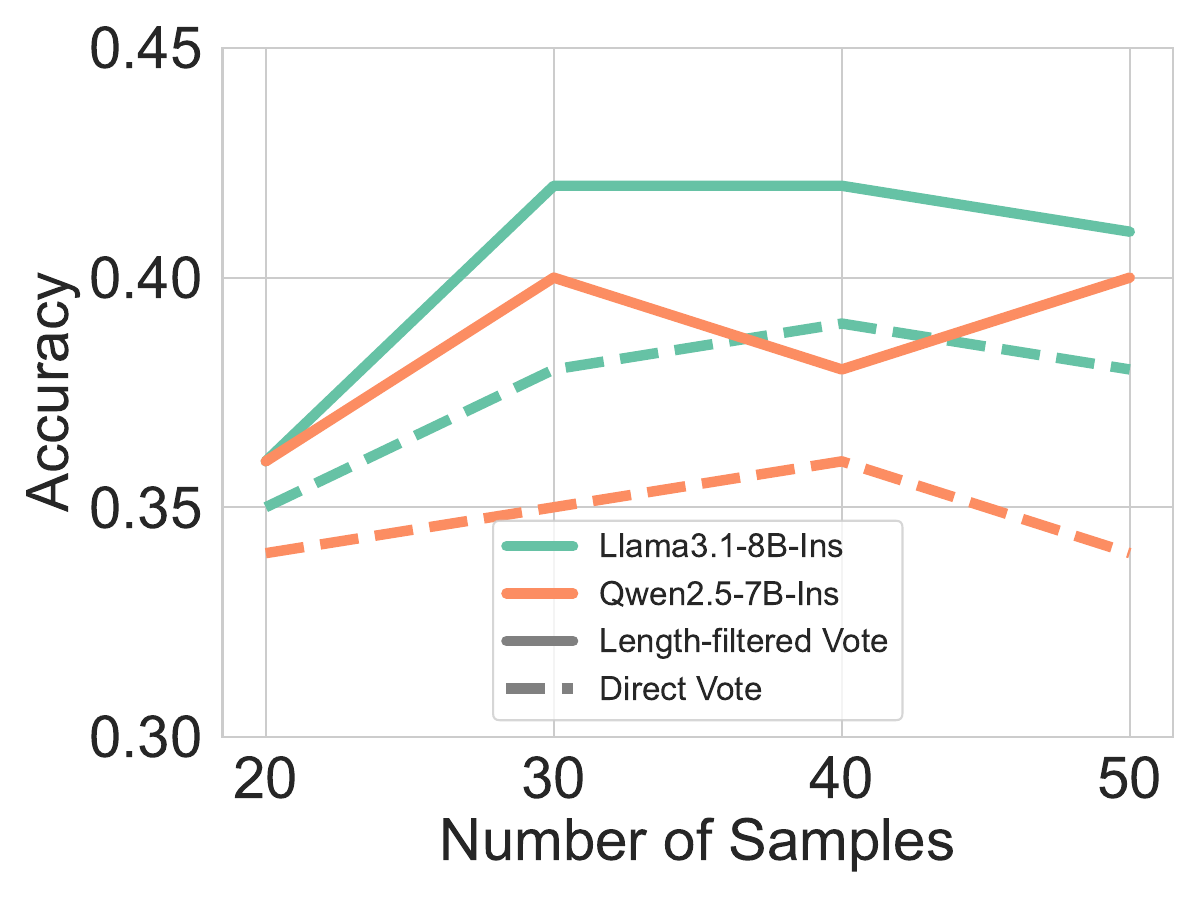}
    \caption{Length-Filtered Vote}
    \label{fig:length-vote}
\end{subfigure}
\caption{(a) and (b) compare model performance under different pretraining data distributions: Mixed Length (uniform over all lengths) vs. Optimal Length (only optimal-length solutions). Despite its smaller size, the small (6 layer) model trained on optimal-length data outperforms the large (9 layer) model trained on mixed-length data, with the performance gap widening as task difficulty increases.
(c) Our Length-Filtered Vote method consistently outperforms vanilla majority vote on the GPQA dataset, maintaining strong performance even as the number of samples increases.}
\label{figure:applications} 
\end{figure*}

\textbf{Training with Optimal-Length CoT Data:} The existence of an adaptive, optimal CoT length suggests that one should design the CoT training data adaptively to fully optimize the model's reasoning performance. To examine the influence of the CoT length of the training data, 
% is  towards optimal, often shorter, CoTs appears to be a phenomenon that models can cultivate for the training process.
% Given that an optimal CoT length exists for specific model capabilities and task difficulties (Figure~\ref{fig:heatmap_synthetic}), a natural question is whether training models predominantly on such optimal-length data can improve performance. We 
we train a model on a specialized dataset that contains
CoT solutions with lengths known to be optimal for the given model size and task difficulty ($T$). We compare this model against a baseline model trained on a dataset of CoT solutions with uniformly distributed step lengths $t$. During testing, models were allowed to freely choose their CoT strategy. 

\textbf{Results.} As shown in Figures~\ref{fig:opt_vs_rand_32} and \ref{fig:opt_vs_rand_64}, the model trained on optimal-length CoTs significantly outperforms the models trained on mixed-length solutions. Remarkably, a smaller model (e.g., 6 layers) trained on optimal-length data can even outperform a larger model (e.g., 9 layers) trained on randomly chosen CoT lengths. This  proof-of-concept experiment  underscores the  critical influence of the suitability of the CoT length in training data for the model. While it is generally hard to exactly estimate optimal CoT lengths in real-world problems, our theoretical and empirical studies provide valuable guidelines for a coarse estimate. We leave more in-depth studies to future work.

\subsection{Adaptive Length-Filtered Vote at  Inference Time}
\label{sec:length_filtered_vote}

The observation that CoTs of optimal length yield higher accuracy suggests that inference-time strategies could benefit from this insight. Standard approaches like majority voting over multiple sampled CoTs, such as self-consistency \citep{self-consistency}, treat all valid reasoning paths equally, regardless of their length. However, paths that are too short (underthinking) or too long (overthinking and error-prone) may contribute noisy or incorrect answers to the voting pool.

Inspired by our findings, we propose \textbf{Length-Filtered Vote}, an adaptive method that enhances standard majority voting by preferentially weighting or exclusively considering answers derived from CoTs whose lengths fall within a proper range. Specifically, in majority vote, given a model \( f_\theta \), a question \( q \), a ground truth answer \( a^* \), we first sample a set of answer candidates $c_1, \dots, c_n \stackrel{i.i.d.}{\sim} f_\theta(q)$ independently. After that, instead of a direct vote, we group the answers by their corresponding CoT length $\ell(c_i)$ (discussed in Appendix~\ref{app:real_world}) into groups with equal bin size $D$ (by default, we set $D=2$), denoted as $\{L_j\}_{j=1}^m$. 
% \sj{what is ``bandwidth'' here? bin size?} 
    As our theory suggests that the prediction accuracy is peaked around a certain range of CoT length, we identify such groups through the prediction uncertainty of the answers within each group, based on the intuition that lower uncertainty implies better predictions. Specifically, we calculate the Shannon entropy $H(L_i)$ of the final answers given by the CoT chains in each group $L_i$. We use a majority vote only for the $K$ (by default, we set $K=3$) groups with the smallest entropy. A detailed description of the algorithm is in Appendix~\ref{app:length_filtered_vote}.

%then select $K$ (by default, we set $K=3$) of the $M$ groups with the smallest entropy and perform a majority vote only on these selected groups. We summarize it in Algorithm \ref{alg:la-vote-top-k}.

\textbf{Results.} We evaluate the proposed method against vanilla majority vote (i.e., self-consistency \citep{self-consistency}) on a randomly chosen subset of 100 questions from the GPQA dataset \citep{gpqa}, a more challenging collection of multiple-choice questions. The results in Figure~\ref{fig:length-vote} show that our filtered vote consistently outperforms vanilla majority vote at different sample numbers and shows little performance degradation as the sample number increases. This further underlines the importance of considering CoT length in the reasoning process.

\section{Related Work}
\label{sec:related_work}

\textbf{Chain-of-Thought for LLM Reasoning.}
% Large Language Models (LLMs) \citep{icl:Brown}
CoT has become a core technique for LLMs to solve
complex reasoning tasks by generating intermediate steps \citep{cot:Wei}. Numerous variants arise to enhance CoT reasoning with more structural substeps, such as least-to-most prompting \citep{least-to-most}, tree of Thoughts \citep{tree-of-thoughts}, and divide-and-conquer methods \citep{Divide-and-Conquer,dacMultichoice}.
% split inputs for more manageable processing. 
These methods fundamentally treat CoT as a framework for task decomposition and subtask solving that falls in our analysis in Section~\ref{sec:theoretical_analysis}.
% , a perspective shared by our study which focuses on the critical aspect of CoT length.

% \textbf{CoT Understanding}
% Many works aim to formalize why CoT is effective. Circuit complexity theory has analyzed the computational power of transformers with and without CoT \citep{cot-theory:wlw,cot-theory:tcs}. \citet{cot-theory:icl} showed that coherent CoT (integrating earlier steps) enhances error correction. Information theory has been used to quantify information gain per step \citep{cot-theory:infor-theory}. Gradient analysis has shown CoT leads to more stable learning \citep{cot-theory:gradient}. \citet{cot-theory:allen-zhu} investigated CoT in synthetic settings to uncover mechanisms. Our work differs by focusing on the performance implications of varying CoT *lengths* and the resulting trade-offs.

\textbf{Overthinking in CoT Reasoning.}
% \yw{mention all concurrent overthinking works}
With the rise of powerful reasoning models like OpenAI o1, scaling test-time compute with long CoT has gained prominence \citep{test-time:0,test-time:1,test-time:2,test-time:3}. These studies often suggest that more computation like longer CoT can lead to better results. However, this is not always true. 
% \citet{are-more-calls-you-need} found that Best-of-N performance can decline if $N$ is too large. 
With similar interests as ours, a few concurrent works also investigated the ``overthinking'' phenomenon \citep{2+3} where reasoning models generate excessively long CoTs for simple problems and proposed some mitigation strategies~\cite{han2024token,luo2025o1,ma2025cot,sui2025stop}. Our analysis not only reveals the inverted-U curve of CoT length and the existence of optimal CoT length, but also  provides a in-depth understanding on
the scaling behaviors and simplicity bias of the optimal CoT length, as supported by both controlled experiments and theoretical analysis. 
This establishes a systematic explanation of overthinking and points out principled guidelines for better CoT designs.
% In this way, our analysis provides a clear
% We also challenge common understanding of RL training by revealing the simplicity bias of RL.
% showing that CoT  
% Our work directly addresses this balance by characterizing the inverted U-shaped performance curve with CoT length and identifying an optimal length. We provide a theoretical basis for why neither too short (underthinking) nor too long (overthinking leading to error accumulation) CoTs are ideal, and how this optimum shifts with model capability and task difficulty.

\textbf{Simplicity Bias and Occam's Razor in Machine Learning.}
The simplicity bias of CoT identified in our work resonates with broader principles like Occam's Razor, which favors simpler explanations or models. In machine learning, a 'simplicity bias' often refers to neural networks learning simpler functions first or being biased towards solutions with lower intrinsic complexity \citep{arpit2017closer,huh2021low}. Our findings extend this understanding to the realm of generated reasoning paths: we reveal that even structured, multi-step reasoning processes like CoT, as produced by LLMs, exhibit such simplicity bias by favoring concise reasoning paths, particularly as model capability increases.

% This is manifested in how models, or the training dynamics that shape them, favor more concise yet effective reasoning paths during their explicit generation, or 'rollouts', 
% \textbf{Simplicity and Occam's Razor in Machine Learning.}
% The observed simplicity bias of CoT length in our work resonates with broader principles like Occam's Razor, which favors simpler explanations or models. In machine learning, simplicity bias often refers to neural networks learning simpler functions first or being biased towards solutions with lower complexity \citep{arpit2017closer,huh2021low}. Our discovery of the simplicity bias of CoT extends previous understanding by revealing that neural-symbolic reasoning like CoTs also exhibits such simplicity bias during explicit rollouts.

% \sj{the need for a model-specific CoT could also be related to algorithmic alignment...}

\textbf{Theoretical Understanding of CoT.} Numerous studies aim to theoretically formalize the Chain-of-Thought (CoT) process and understand its effectiveness. They include analyzing CoT's computational advantages via circuit complexity \citep{cot-theory:wlw,cot-theory:tcs}, demonstrating how coherent reasoning paths enhance error correction and accuracy \citep{cot-theory:icl}, and quantifying step-wise information gain from an information-theoretic standpoint \citep{cot-theory:infor-theory}. Further research has shown that detailed CoT improves learning stability by affecting gradient dynamics \citep{cot-theory:gradient}, while controlled synthetic experiments have helped uncover underlying problem-solving mechanisms in LLMs \citep{cot-theory:allen-zhu}. Distinct from these varied theoretical explorations, our theory characterizes how CoT length influences final performance and explains its scaling behaviors  through the interplay of task decomposition and error accumulation. Furthermore, our findings on CoT scaling behaviors and the consequent need for model-specific CoT structures (as discussed in Section~\ref{sec:practical_insights_synthetic}) resonate with the concept of algorithmic alignment \citep{xu2019can}, which suggests that models perform best when the problem structure aligns with their computation structure.

\section{Conclusion}
\label{sec:conclusion}
In this paper, we challenged the notion that longer Chain-of-Thought (CoT) processes are invariably superior, demonstrating through extensive experiments and theoretical analysis that CoT length and accuracy typically follow an inverted U-shaped curve, implying an optimal length that balances task decomposition against error accumulation. We discovered the simplicity bias of CoT, where more capable models prefer shorter effective reasoning paths, and formally derived scaling laws for this optimal length relative to model capability and task difficulty. Practically, we showed that reinforcement learning can guide models towards this optimal CoT length, that training on optimally-lengthed CoTs boosts performance, and proposed "Length-Filtered Vote" as a promising inference strategy. Our work underscores the critical need to calibrate CoT length, moving beyond a one-size-fits-all approach towards a principled framework where LLMs adaptively choose the right amount of thought to optimize reasoning.

\section*{Acknowledgement}
Yisen Wang was supported by National Key R\&D Program of China (2022ZD0160300), National Natural Science Foundation of China (92370129, 62376010), and Beijing Nova Program (20230484344, 20240484642).
Yifei Wang and Stefanie Jegelka were supported in part by the NSF AI Institute TILOS (NSF CCF-2112665), and an Alexander von Humboldt Professorship.

\bibliography{ref}
\bibliographystyle{plainnat}

\newpage
\appendix
\onecolumn

\begin{center}
\LARGE \textbf{Appendix}
\end{center}

\etocdepthtag.toc{mtappendix}
\etocsettagdepth{mtchapter}{none}
\etocsettagdepth{mtappendix}{subsection} % Show subsections in appendix ToC
\tableofcontents

%yuyang wu: This is the appendix for nips version
%The content of appendix should be A. Problem Formulation B. Real world details C. synthetic details D Theory

\section{Limitations}
\label{app:limitations}
This proof-of-concept study highlights the critical role of aligning Chain-of-Thought (CoT) length with model capability and task difficulty, particularly in training data. However, accurately estimating the optimal CoT length in real-world scenarios remains challenging due to the complexity and variability of reasoning tasks. While our theoretical and empirical analyses offer practical heuristics for coarse approximation, they may not fully capture the nuances of diverse problem domains or model behaviors. We leave the development of more precise estimation methods and adaptive strategies for optimal CoT length selection in complex, real-world settings as promising directions for future work.

\section{Formal Definitions of Simplified Arithmetic Problem}
\label{app:setup}
\begin{wrapfigure}{r}{0.5\linewidth}
  \centering
  \begin{tikzpicture}
    [scale=1, level distance=0.75cm, sibling distance=1.5cm,
     every node/.style={draw, circle, minimum size=0.5cm, inner sep=1.5pt}]
     
    \node {+}
      child { node {5} }
      child { node {+}
        child { node {4} }
        child { node {+} 
          child { node {+} 
            child { node {2} }
            child { node {1} }
          }
          child { node {3} }
        }
      };
  \end{tikzpicture}
  \caption{Computation tree of arithmetic expression $5+(4+((2+1)+3))$.}
  \label{fig:tree}
\end{wrapfigure}
To begin, we aim to empirically investigate the relationship between reasoning performance and CoT length. Therefore, we need to control a given model to generate reasoning chains of varying lengths for a specific task. Unfortunately, no existing real-world dataset or model fully meets these strict requirements. Real-world reasoning tasks, such as GSM8K or MATH \citep{gsm8k, math}, do not provide multiple solution paths of different lengths, and manually constructing such variations is challenging. Moreover, it is difficult to enforce a real-world model to generate a diverse range of reasoning paths for a given question.  
Given these limitations, we begin our study with experiments on synthetic datasets.

\subsection{Problem Formulation}

To investigate the effect of CoT length in a controlled manner, we design a synthetic dataset of simplified arithmetic tasks with varying numbers of reasoning steps in the CoT solutions.
\begin{definition}[Problem]
\label{def:task}
    In a simplified setting, an arithmetic task $q$ is defined as a binary tree of depth \( T \). The root and all non-leaf nodes are labeled with the \( + \) operator, while each leaf node contains a numerical value (mod 10). In addition, we impose a constraint that every non-leaf node must have at least one numerical leaf as a child. 
\end{definition}

The bidirectional conversion method between arithmetic expressions and computation trees is as follows: \textit{keeping the left-to-right order of numbers unchanged, the computation order of each "+" or tree node is represented by tree structure or bracket structures.}
For example, consider the task \( 5+(4+((2+1)+3)) \) with $T=4$. The corresponding computation tree is defined as Figure~\ref{fig:tree}.

To ensure that CoT solutions of the same length have equal difficulty for a specific problem, we assume that each reasoning step performs the same operations within a single CoT process.
\begin{definition}[Solution]
\label{def:solu}
    We define a $t$-hop CoT with a fixed each step length of \( t \) as a process that executes \( t \) operations starting from the deepest level and moving upward recursively. 
\end{definition}

According to this definition, the execution sequence is uniquely determined. For example, one way to solve expression in Figure~\ref{fig:tree} is by performing one addition at a time:
\begin{align}
    &5+(4+((2+1)+3)) = \texttt{<1>}\\ \label{eq:subtask}
    &2 + 1 = 3\\ \notag
    &3 + 3 = 6\\ \notag
    &4 + 6 = 0\\ \notag
    &5 + 0 = 5 \texttt{<END>.} \notag
\end{align}
Another approach is to perform two additions at a time:
\begin{align}
    &5+(4+((2+1)+3)) = \texttt{<2>}\\ \notag
    &(2 + 1 ) + 3  = 6\\ \notag
    &5 + (4 + 6) = 5\texttt{<END>.} \notag
\end{align}
The latter approach is half as long as the former, but each reasoning step is more complex\footnote{This is because performing two operations at once requires the model to either memorize all combinations of numbers in a two-operator equation and their answers, apply techniques like commutativity  to reduce memory requirements, or use its mental reasoning abilities to perform the two operations without relying on CoT.}. This illustrates a clear trade-off between the difficulty of each subtask and the total number of reasoning steps. 

In practice, when \( t \) does not evenly divide \( T \), the final step performs \( T \mod t \) operations. To guide the model in generating the desired CoT length, we insert the control token \texttt{<t>} after the question and before the beginning of the solution. To preserve the parentheses that indicate the order of operations, we construct expressions in Polish notation. However, for readability, we present each problem in its conventional form throughout the article.

\subsection{Contrast to vanilla arithmetic problem}
\textbf{Why pruning?}
Initially, we intended to create a synthetic dataset for regular arithmetic tasks, but we quickly realized that the computation tree for such tasks is uncontrollable. For example, consider the task $1*2 + 3*4$. We hoped to compute $2$ operators in one step, but found it impossible because the addition needs to be computed after the two multiplications, and we cannot aggregate two multiplications in one subtask. Therefore, pruning the computation tree becomes essential.

\textbf{Why only focusing on addition?}
There are two reasons why we focus on arithmetic tasks involving only addition: first, it simplifies pruning, as the order of operations can be controlled solely by parentheses; second, it facilitates the computation of sub-tasks, since parentheses do not affect the final result, and the model only needs to compute the sum of all the numbers when solving a sub-task. We aim for the model to handle longer sub-tasks, thereby allowing a broader study of the impact of CoT length.

\textbf{Will the simplified synthetic dataset impact the diversity of the data?}
We need to clarify that even with pruning, the structure of the expressions will still vary because swapping the left and right child nodes of each non-leaf node in the computation tree results in different expressions. When $T > 30$, the number of possible variations exceeds $1 \times 10^9$.

\section{Supplementary Details on Real world Experiment for Optimal CoT Length}
\label{app:real_world}
\subsection{Implementation Details}
% \wyy{Here, we need to provide experiment statistical significance}
\textbf{Solution Length Control.}
To study the impact of CoT length on performance under a given problem difficulty, we need to induce the model to naturally generate solutions of varying lengths. Simply adding prompts like “\textit{please use 100 tokens to solve this problem}” or “\textit{please use 10 steps to solve this problem}” is not ideal because the model’s ability to follow instructions regarding output length is limited, and such fixed-length prompts may not ensure fairness across problems of different difficulties. Moreover, prompting for a specific length might lead the model to generate irrelevant tokens or steps just to “pad the length,” without actually changing the number of steps or the complexity of the reasoning. Additionally, controlling \texttt{max\_length} is also problematic, as overly long responses might get truncated, which would directly lead to lower accuracy for longer outputs. What we really want is for the model to generate a complete and coherent long response on its own, so we can observe the corresponding accuracy.

To create solutions with varying step lengths with different complexity, we follow \citep{complex-prompt} by using in-context examples (8-shots) with three different levels of complexity to guide the model in generating solutions with different step counts. For each set of in-context examples, we sample 20 times, resulting in a total of 60 samples per question. 

\textbf{Step Segmentation.}
Simply measuring CoT length by counting tokens is neither rigorous nor meaningful. Since our focus is on final performance rather than efficiency, we care more about using CoT length to reflect the complexity of the reasoning pattern. In this sense, the number of reasoning steps can serve as a more appropriate indicator of CoT length. As we discussed earlier, the step number captures how the model decomposes the problem, which directly reflects the complexity of its reasoning. In contrast, token length fails to capture this because, as the model thinks more deeply and the number of steps increases, the number of tokens per step may decrease—making the total token count unpredictable and unreliable as a proxy for reasoning complexity.

When calculating the number of steps, we separate the full reasoning chain using \texttt{"\textbackslash n"}\citep{complex-prompt} and remove empty lines caused by \texttt{"\textbackslash n\textbackslash n"}. Then we consider the total number of lines as the CoT length. Since questions in the MATH dataset are challenging and lead to high variability in final CoT lengths, we normalize the lengths by applying \texttt{length = length // bin\_width}. For experiments comparing different models (e.g., optimal CoT length per model or optimal vs. longest CoT), the questions within each length bin differ (though only 30 per group), which introduces variability. To reduce this variance and ensure each bin has enough samples, we use a relatively large bin width of 5. In contrast, for analyzing the influence of task difficulty, where each calculation on optimal CoT length only contains one question, we adopt a finer bin width of 2 for better resolution (we also verified that using width 1 yields almost identical results).

\textbf{More Details of Figure~\ref{fig:real_world_task_difficulty_vs_opt_length}.} When evaluating the results, questions with accuracy $<0.01$ or $>0.99$ (indicating all incorrect or all correct responses) are excluded, as their accuracy does not vary with step length changes.

To better understand the reliability of the observed trend between task difficulty and optimal Chain-of-Thought (CoT) length, we compute a 95\% confidence interval around the linear regression line. Specifically, we use standard methods based on the Student’s t-distribution to estimate uncertainty in the predicted values. The confidence band reflects how much the estimated mean CoT length is expected to vary given the finite sample size and the distribution of data points.
\subsection{More Experimental Results}
\label{app:difficulty}

\begin{figure*}[ht]
\centering

\begin{subfigure}{0.45\textwidth}
    \centering
    \includegraphics[width=\linewidth]{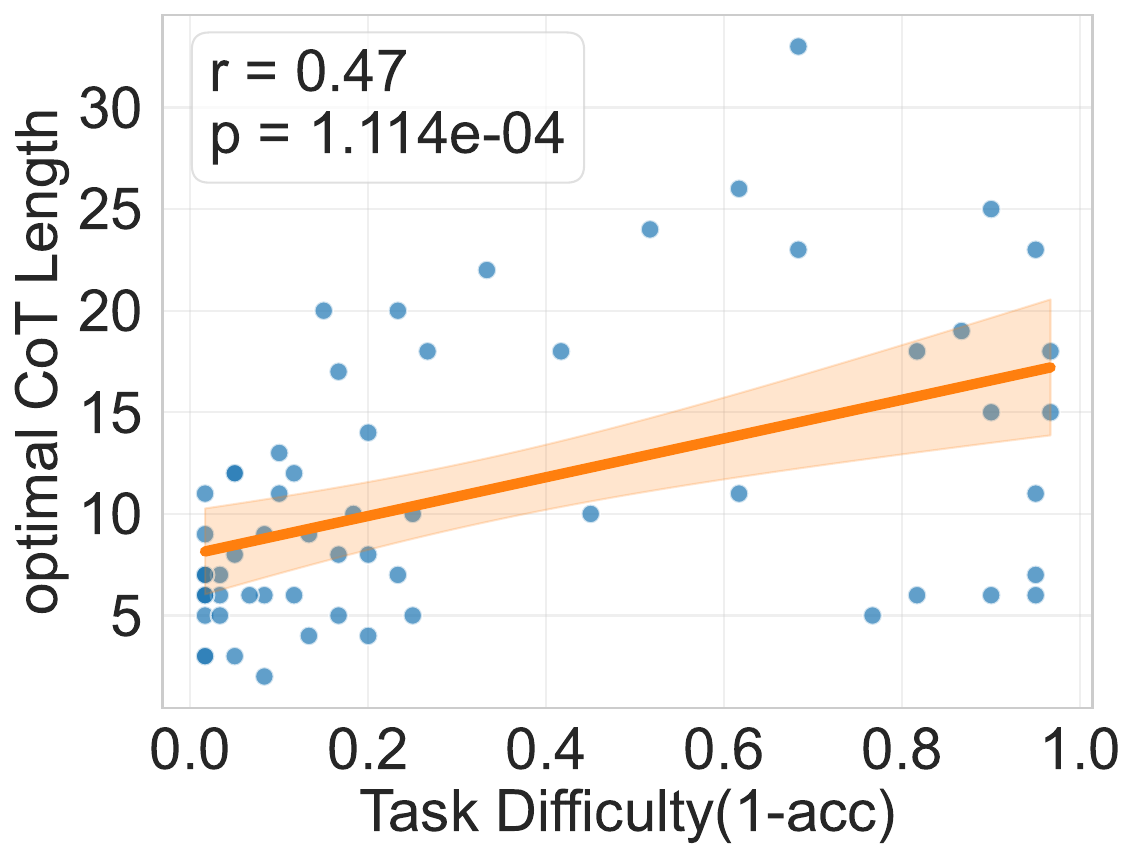}
    \caption{Qwen2.5 7B Instruct}
    \label{fig:step_vs_acc_qwen7b}
\end{subfigure}%
% \hspace{0.03\textwidth}  % Adjusted spacing between figures
\begin{subfigure}{0.45\textwidth}
    \centering
    \includegraphics[width=\linewidth]{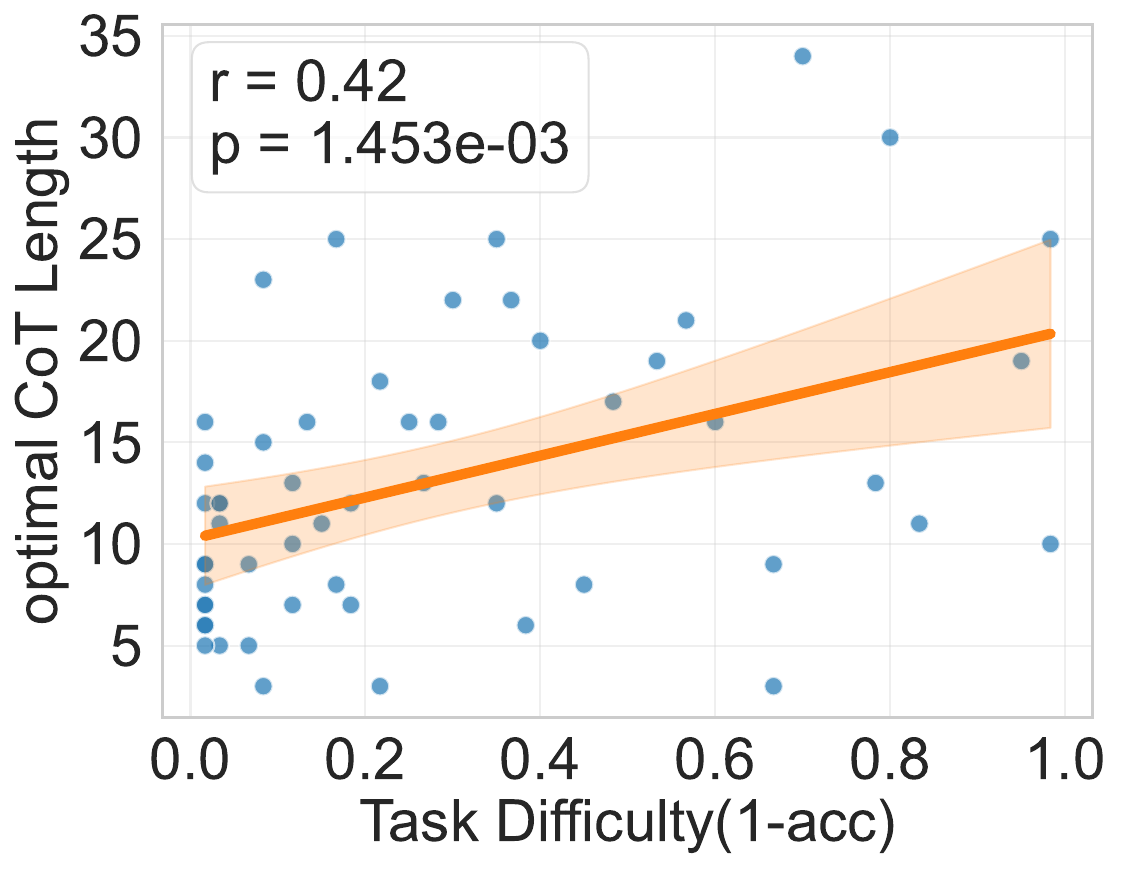}
    \caption{Qwen2.5 14B Instruct}
    \label{fig:step_vs_acc_qwen14b}
\end{subfigure}
\caption{Evaluation between task difficulties and optimal CoT lengths on MATH datasets with Qwen2.5 Series Instruct models. }
\label{fig:qwen}
\end{figure*}

\begin{figure*}[ht]
\centering

\begin{subfigure}{0.45\textwidth}
    \centering
    \includegraphics[width=\linewidth]{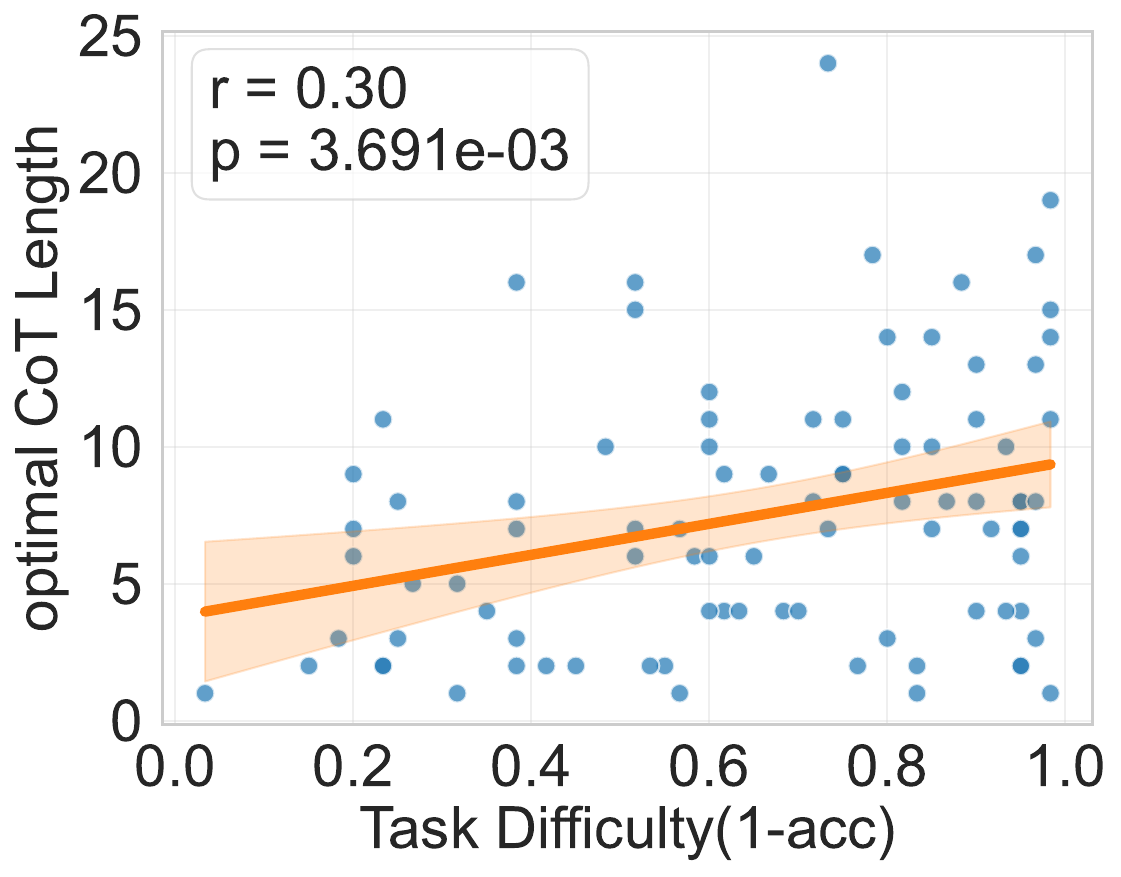}
    \caption{Llama 3.1 8B Instruct}
    \label{fig:step_vs_acc_llama8b}
\end{subfigure}%
% \hspace{0.03\textwidth}  % Adjusted spacing between figures
\begin{subfigure}{0.45\textwidth}
    \centering
    \includegraphics[width=\linewidth]{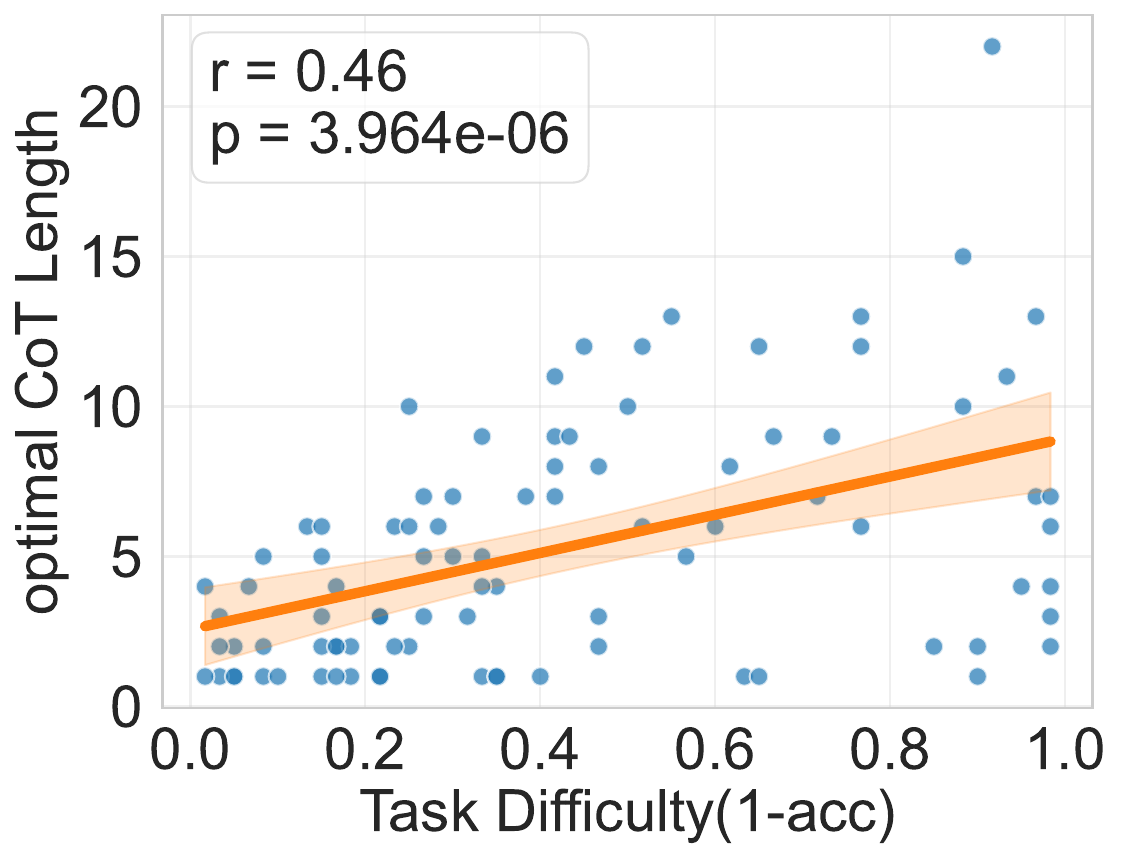}
    \caption{Llama 3.1 70B Instruct}
    \label{fig:step_vs_acc_llama70b}
\end{subfigure}
\caption{Evaluation between task difficulties and optimal CoT lengths on MATH datasets with LLama3.1 Series Instruct models. 
}
\label{fig:llama}
\end{figure*}

\textbf{Additional results for Figure~\ref{fig:real_world_task_difficulty_vs_opt_length}.} To further investigate the relationship between task difficulty and optimal CoT lengths on real world datasets, we conduct experiments on different models. The results (Figure \ref{fig:qwen} and \ref{fig:llama}) are impressive that results on all models show a significant correlation between the task difficulties and optimal lengths.

\begin{figure}[H]
\begin{subfigure}{0.49\textwidth}
    \centering
    \includegraphics[width=\linewidth]{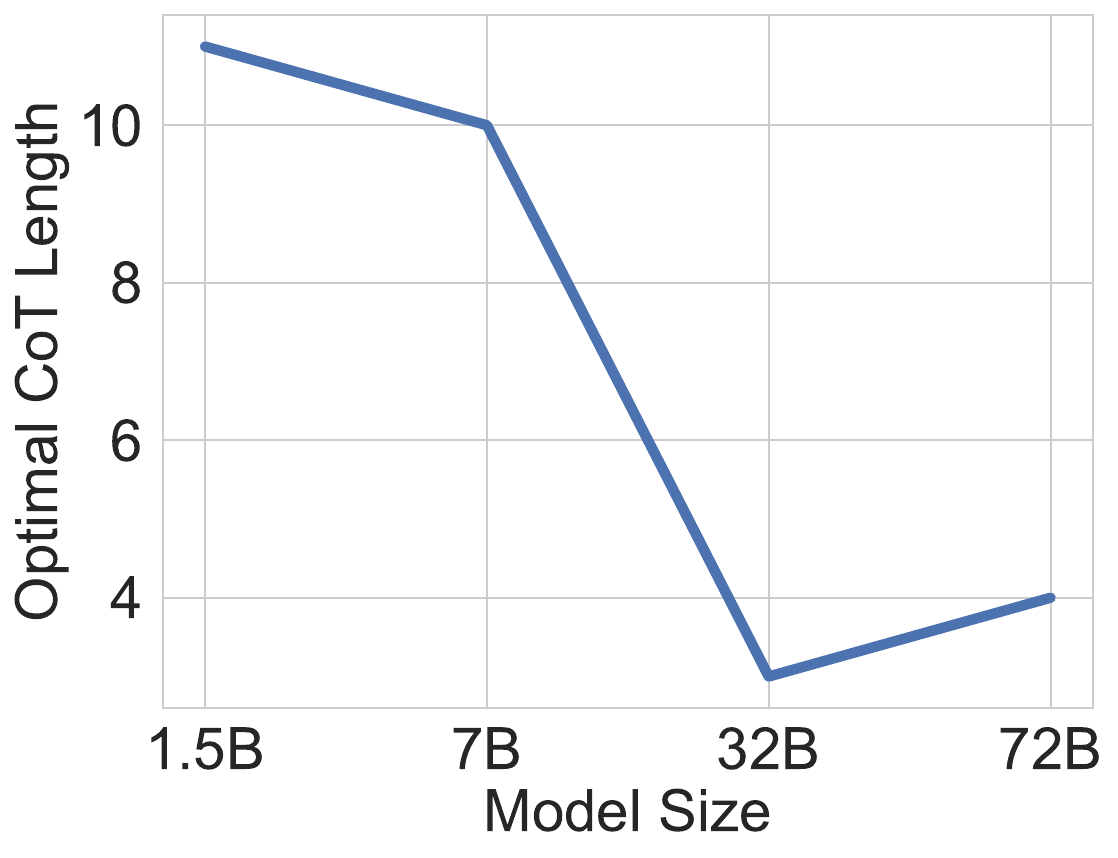}
    \caption{Optimal CoT length vs. Model size (Qwen2.5 series).}
    \label{fig:real_world_model_size_vs_opt_length_mmlu}
\end{subfigure}%
\hfill
\begin{subfigure}{0.49\textwidth}
    \centering
    \includegraphics[width=\linewidth]{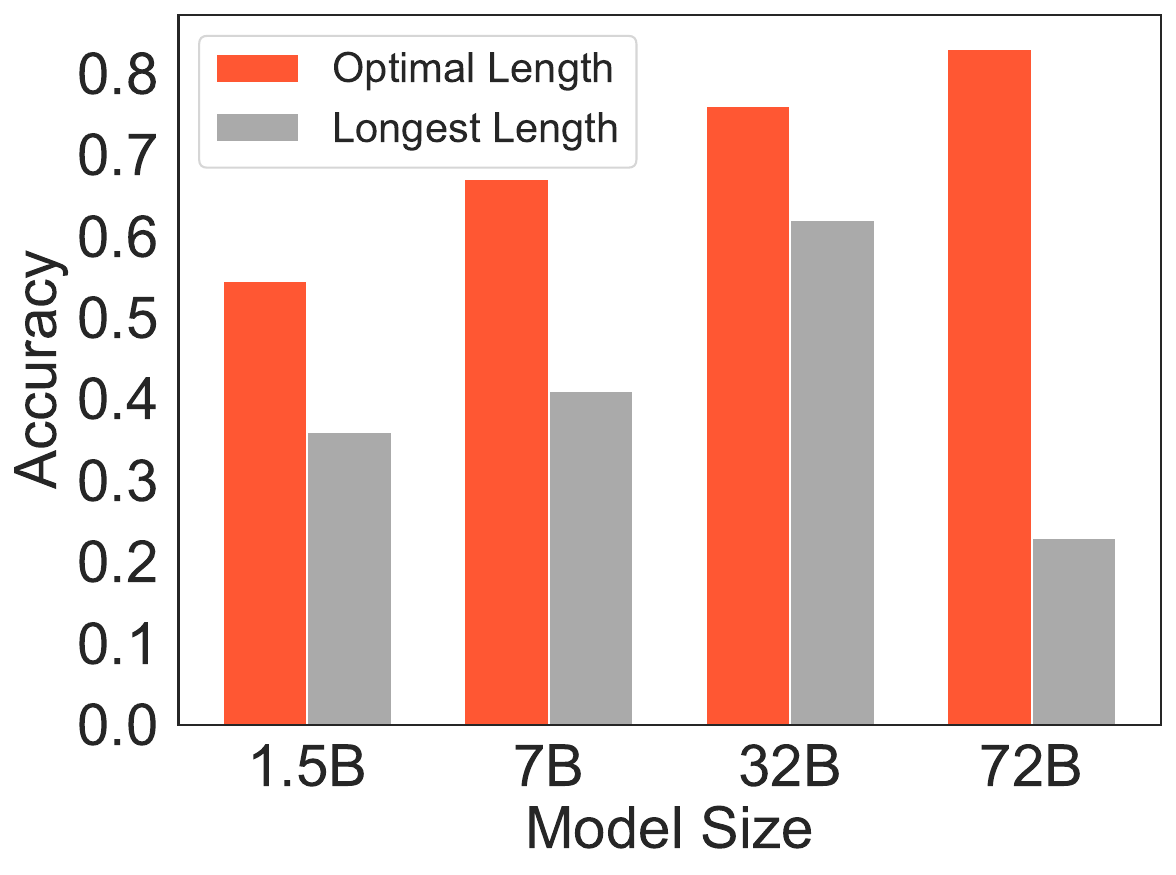}
    \caption{Optimal vs. Longest CoT length accuracy on MMLU STEM dataset.}
    \label{fig:real_world_acc_gap_mmlu}
\end{subfigure}%
\caption{Real-world CoT length observations. (a) Larger models tend to achieve optimal performance with shorter CoTs. (b) Accuracy for CoTs of optimal length is significantly higher than that of the longest CoTs.}
\label{fig:real_world_observations_mmlu}
\end{figure}

\begin{figure}[H]
\vspace{-20pt}
    \centering
    \begin{subfigure}{0.49\textwidth}
        \centering
        \includegraphics[width=\linewidth]{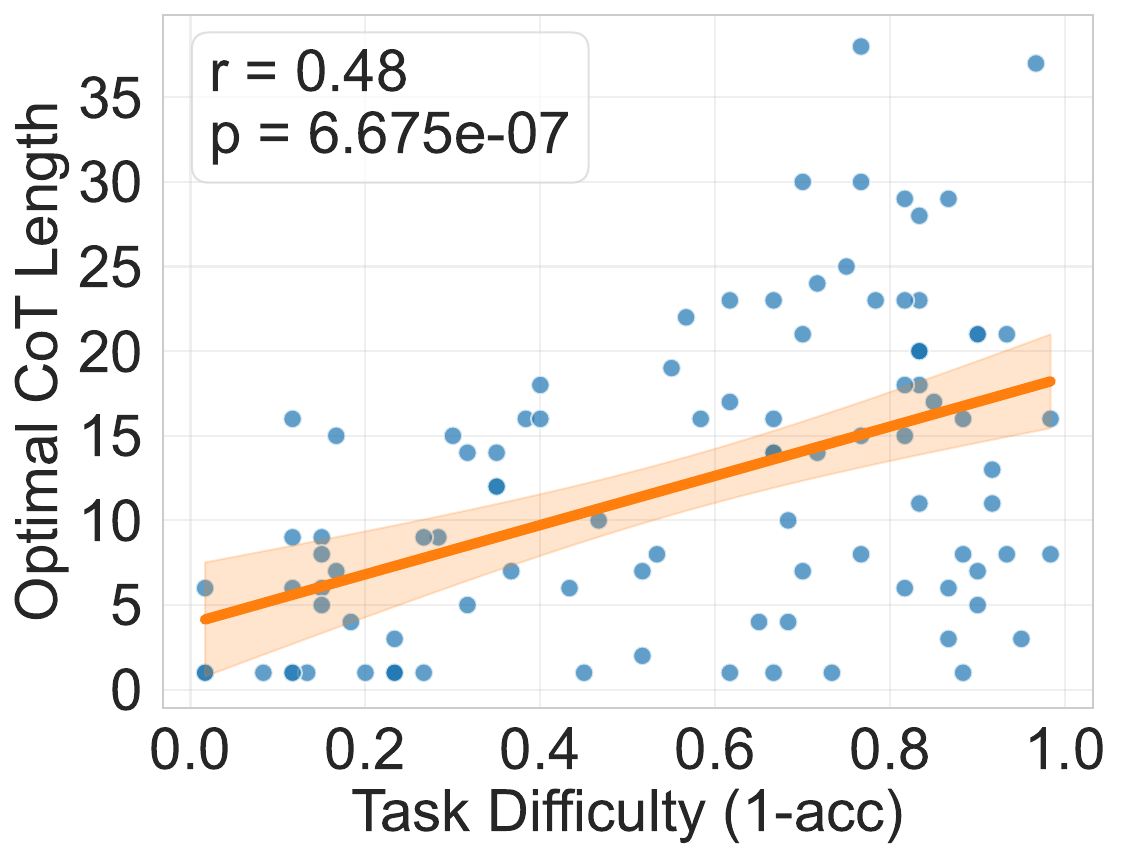}
        \caption{Qwen2.5-1.5B-Instruct}
        \label{fig:mmlu_1.5}
    \end{subfigure}
    \hfill
    \begin{subfigure}{0.49\textwidth}
        \centering
        \includegraphics[width=\linewidth]{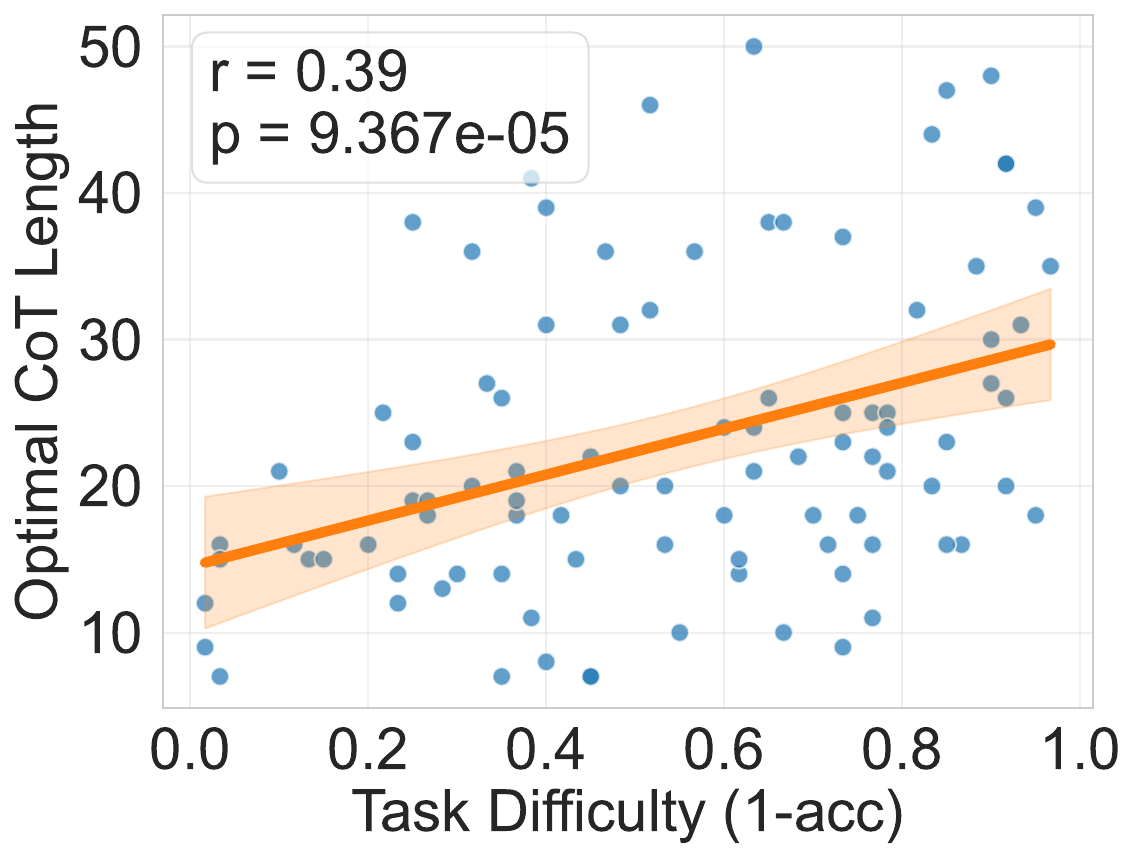}
        \caption{Qwen2.5-7B-Instruct}
        \label{fig:sub2}
    \end{subfigure}

    \vspace{0.5cm} % 控制上下图间距

    \begin{subfigure}{0.49\textwidth}
        \centering
        \includegraphics[width=\linewidth]{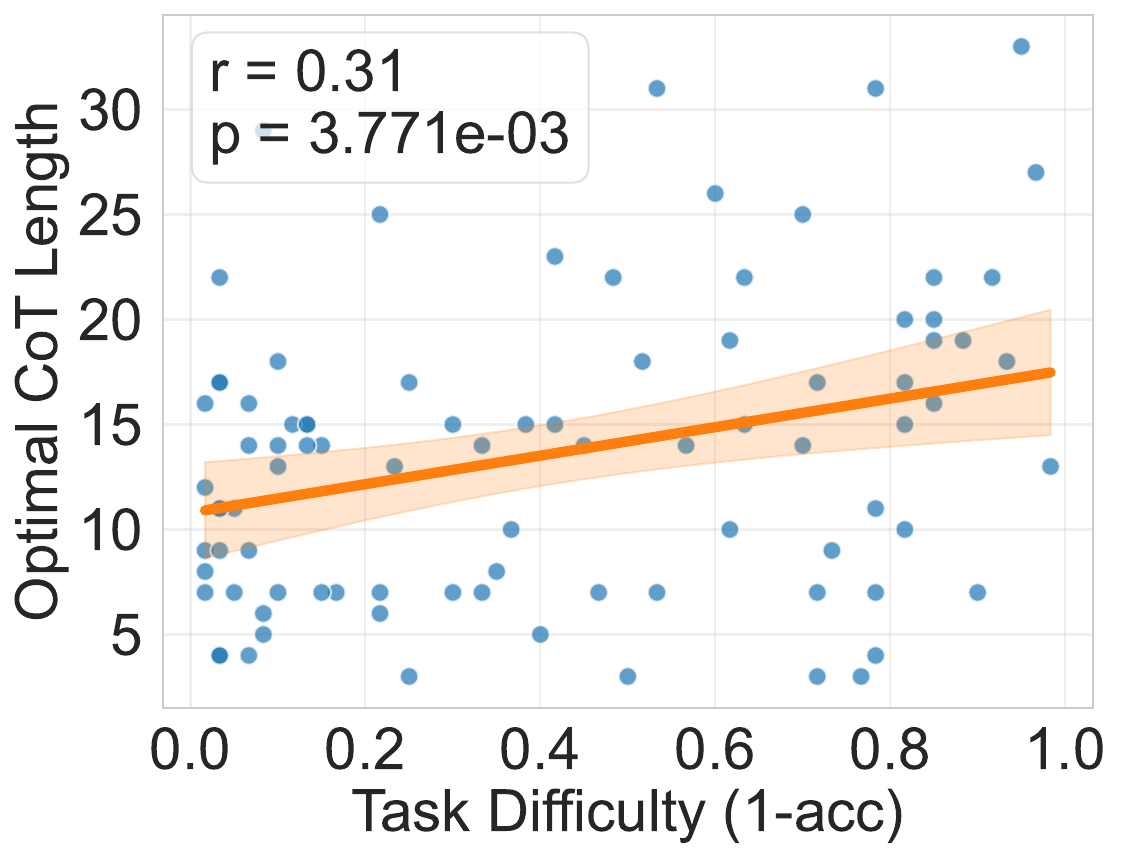}
        \caption{Qwen2.5-32B-Instruct}
        \label{fig:mmlu_32}
    \end{subfigure}
    \hfill
    \begin{subfigure}{0.49\textwidth}
        \centering
        \includegraphics[width=\linewidth]{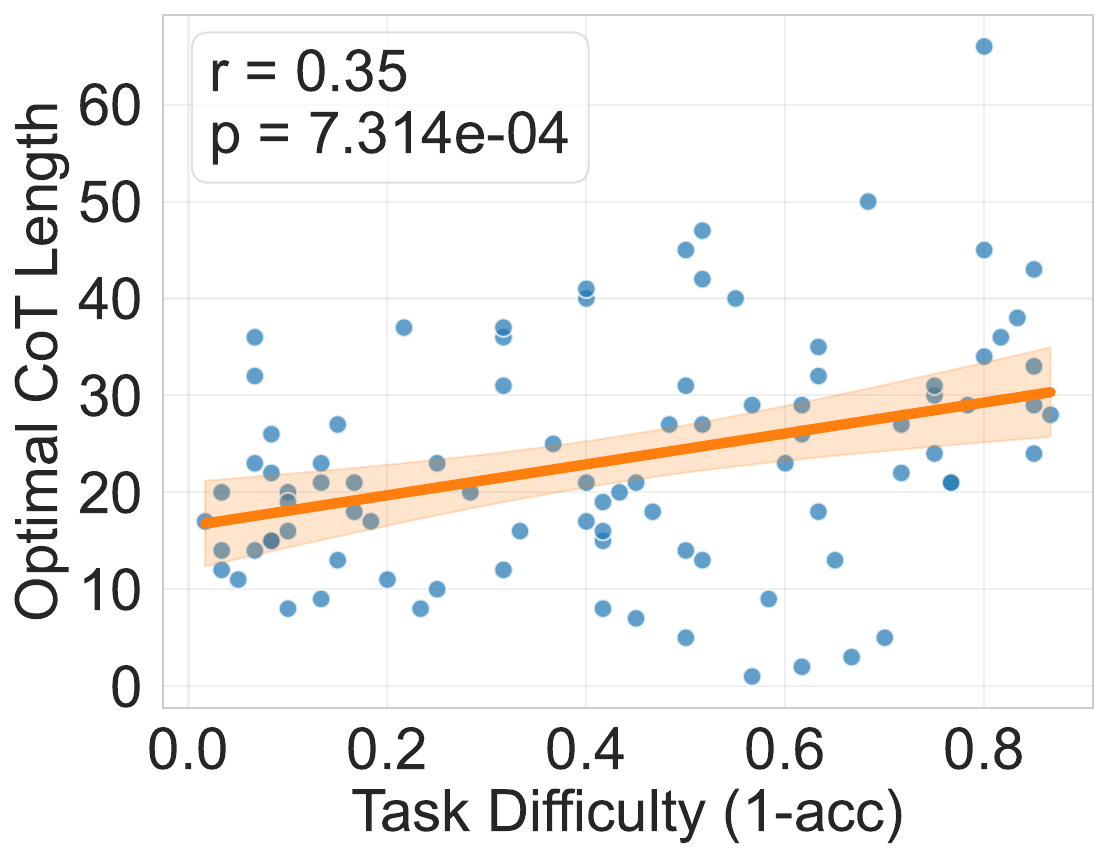}
        \caption{Qwen2.5-72B-Instruct}
        \label{fig:mmlu_72}
    \end{subfigure}
    \caption{Evaluation between task difficulties and optimal CoT lengths on MMLU STEM datasets with
Qwen2.5 Series Instruct models.}
    \label{fig:2x2}
    
\end{figure}

\textbf{Results on MMLU STEM dataset.} We also conduct experiments on the MMLU STEM dataset using the Qwen2.5 Series instruct models under the same settings as the MATH dataset. The results, shown in Figures~\ref{fig:real_world_observations_mmlu} and~\ref{fig:2x2}, exhibit similar trends to those observed on the MATH dataset.

\section{Supplementary Details on Real World Experiment for RL Simplicity Bias}\label{app:exp-real}
% \wyy{@ZIYU you can put experimental detail and more results here.}

For Figure~\ref{fig:real_world_rl_cot_decrease_conceptual}, we use Qwen2.5-7B-Instruct~\citep{qwen} as the base model, Group Relative Policy Optimization with R1-like prompting~\citep{grpo,r1} for the reinforcement learning process, and LeetCode-2K~\citep{leetcodedataset} as the training and evaluation dataset. We take the following training configuration by default:
\vspace{-10pt}

\begin{table}[H]
\centering
\caption{Hyperparameter settings for real-world RL experiments with Qwen2.5-instruct models.}
\vspace{+3pt}
\label{tab:my-table}
\resizebox{.96\columnwidth}{!}{%
\begin{tabular}{@{}cccccc@{}}
\toprule
Learning Rate & Max Epochs & Rollout Samples & Reverse KL  Coefficient & Entropy Loss Coefficient & Effective Batch Size \\ \midrule
5e-7          & 10         & 16              & 1e-3               & 5e-3    &   256              \\ \bottomrule
\end{tabular}%
}
\end{table}

\section{Additional Synthetic Experiment Details}
\label{app:syn}
\subsection{Training details}
In default, we train different models (layers ranging from 5 to 9) on the same dataset, which included mixed questions with total operators $T \in [12,80]$ and random sampled CoT solutions with each step operators $t \in [1,12]$. All other parameters are kept the same with the huggingface GPT-2 model. During the training process, the CoT indicator token \texttt{<t>} is also trained, so that during test-time, we can let the model decide which type of CoT it will use by only prompting the model with the question. For each model, we train $25000$ iterations with batch size that equals 256. During test-time, we test 100 questions for each $T$ and $t$. All experiments can be conducted on one NVIDIA A800 80G GPU. 

\subsection{Observation of subtask loss}
\label{app:subtask}
As we observed in training losses, the loss of subtask generation tokens (e.g. $1+2$) for the easiest subtask($t=1$) is about 3 times larger than the hardest subtask ($t=12$), while the loss ratio for subtask answer tokens is $1e4$. Therefore, it is acceptable for taking the subtask error rate constant with $t$.

Besides, there is no obvious pattern showing the model sizes affect the subtask loss. Moreover, the smallest model and the largest model have almost the same subtask loss. Therefore, in our settings, we take model size as irrelevant with the subtask error rate.

%yuyang wu:since we have random error version, it seems that we do not need to specifically discuss this assumption.
% \subsection{Discussion on Same Subtask Difficulty Assumption}
% \label{app:same_length}
% \wyy{discuss in theory sec}
% A Chain-of-Thought (CoT) containing sub-tasks of varying difficulty can be seen as a mixture of CoTs with the same difficulty level. On the other hand, allowing CoTs to generate sub-tasks of different lengths increases the overall difficulty of CoTs of the same length, making them harder to study. Third, under Assumption \ref{ass:error_e}, convexity analysis shows that the final accuracy function (Proposition~\ref{pro: general_acc}) is concave. Therefore, to maximize accuracy, all sub-tasks should have the same difficulty level.

\section{Theoretical Results under Broader Scenarios}
% \label{sec:general_error_theory_revised}
\label{app:extension}
% \subsection{Extension to Broader Scenarios}

\subsection{General Errors}
In the simple case we discussed in Section~\ref{sec:theory}, we discussed the trend of overall accuracy with respect to $N$ and the variation of optimal $N$ with $M$ and $T$, assuming the subtask error rate is a linear function. In the following discussion, we aim to derive conclusions corresponding to more general error rate functions. We find that as long as the error function satisfies some basic assumptions on the \textbf{monotonicity} and \textbf{convexity} of the error functions, the above conclusions still hold.

\label{app:assumption}
\begin{assumption}
     $E(N,M,T)$ satisfies the following reasonable conditions:
\begin{itemize}
\label{ass:error_e}
    \item $0 < E(N=1,M,T) < 1$
    \item $\lim_{N \rightarrow +\infty}E(N,M,T) = 0$
    \item $E(N,M,T)$ is monotonically deceasing with $N$, since more detailed decomposition leads to easier subtask.
    \item $E(N,M,T)$ is convex with $N$, since the benefits of further decomposing an already fine-grained problem($N$ is large) are less than the benefits of decomposing a problem that has not yet been fully broken down($N$ is small).
    \item $E(N,M,T)$ is monotonically deceasing with $M$, since stronger models have less subtask error rate.
    \item $E(N,M,T)$ is monotonically increasing with $T$, since harder total task leads to harder subtask while $N,M$ are the same.
\end{itemize}
\end{assumption}

\begin{assumption}
\label{ass:error_sigma}
     $\sigma(T)$ is monotonically increasing with $T$
\end{assumption}

%  Assuming $0 < \sigma(T) < 1$ and $0 < E(N,M,T) < 1$ satisfy certain reasonable properties (e.g., $E$ decreases with $N$ and $M$, increases with $T$; detailed in , we can state more general results.

With Assumption~\ref{ass:error_e} and \ref{ass:error_sigma}), the core insights from the linear case can be generalized.

\begin{restatable}{theorem}{general}
\label{thm:general_theory}
    For a noise function $0<\sigma(T)<1$ and a subtask error rate function $0<E(N,M,T)<1$ satisfying Assumptions~\ref{ass:error_e} and \ref{ass:error_sigma}, the general final accuracy function $A(N)$ from Proposition~\ref{pro:general_acc} has the following properties:
    \begin{itemize}
        \item $\lim_{N\rightarrow +\infty} A(N) = 0$. (Excessively long chains always fail.)
        \item If $A(N)$ has a maximum at $N^* > 1$, then $N^*$ has a lower bound related to $M$ and $T$:
        \begin{align}
        \label{eq:lower_bound_N_general_revised}
            N^* \geq N_{LB}(M,T) = E_N^{-1}\left(1-\frac{1}{e^2(1-\sigma(T))}; M, T\right),
        \end{align}
        where $E_N^{-1}(\cdot; M, T)$ is the inverse of $E(N,M,T)$ with respect to $N$.
    \end{itemize}
\end{restatable}
The monotonicity of $E_N^{-1}$ with respect to $M$ (decreasing) and $T$ (increasing, assuming $\sigma(T)$ doesn't dominate adversely) implies that the qualitative scaling laws (Corollaries stemming from Theorem~\ref{thm:simple_optimal_N_theory}) still hold under general conditions, supporting the empirically observed Simplicity Bias and the inverted U-shaped performance.

\begin{corollary}
    As the model becomes stronger, $E^{-1}$ decreases monotonically with respect to $M$, which leads to a decrease of $N(M,T)$.
\end{corollary}

\begin{corollary}
    As the task becomes harder, $E^{-1}$ is monotonically increasing with respect to $T$, which leads to an increase in $N(M,T)$.
\end{corollary}

% \subsection{Extension to Random Error Functions}
% \label{sec:random_error_theory_revised}
\subsection{Random Error}
In Theorem~\ref{thm:simple_optimal_N_theory} and~\ref{thm:general_theory}, we make a strong assumption that all sub-question or sub-answer errors are identical, which does not align well with real-world scenarios. In practice, each sub-task may exhibit a different error rate. However, they generally follow a trade-off: the more the task is decomposed, the easier each sub-task becomes. Specifically, we can model the error rate of each sub-task as a random variable with a fixed expectation that monotonically decreases with the number of CoT steps $N$.

To simplify the problem, here we assume $\sigma_i \sim B(\alpha_1(T),\beta_1(T))$ to be the sub-question error rate, and $e_i \sim B(\alpha_2(N,M,T),\beta_2(N,M,T))$ to be the sub-answer error rate. Then, as a variant of Proposition~\ref{pro:general_acc}, the expectation of final accuracy is $\mathbb{E}\left[\prod_{i=1}^N(1-e_i)(1-\sigma_i)\right]$. 

It is worth noting that each $\sigma_i$ or $e_i$ is not independent. If most steps are easy (i.e., have low error rates), the remaining steps are more likely to be easy as well. Moreover, if a particular step serves as a self-validation step, its high accuracy can influence the correctness of other steps that depend on it. This also provides an interpretation for reasoning models exhibiting backtracking behavior.

\begin{restatable}{theorem}{random}
\label{thm:random_error}
Let $\alpha_1 = T$, $\beta_1 = C - T$, $\alpha_2 = T$, and $\beta_2 = NM - T$. Then the expected error rates for sub-questions and sub-answers are given by $\mathbb{E}[\sigma_i] = \frac{T}{C}$ and $\mathbb{E}[e_i] = \frac{T}{MN}$, respectively. Based on these estimates, we can derive an upper bound $\hat{A}(N)$ on the final accuracy 

$$
\mathbb{E}\left[\prod_{i=1}^N (1 - e_i)(1 - \sigma_i)\right] \leq \hat{A}(N) = \left[\left(1 - \frac{T}{C + 2N - 1}\right)\left(1 - \frac{T}{NM + 2N - 1}\right)\right]^N,
$$
which initially increases and then decreases as the number of CoT steps $N$ grows.
\end{restatable}

This suggests that even with stochasticity, the fundamental trade-off leading to an optimal CoT length persists.

% \subsection{Discussion on reasoning model}

\section{Proof}
\label{app:proof}
In this section, we provide the proofs for all theorems.

\subsection{Proof of Proposition~\ref{pro:general_acc}}
\finalacc*
\begin{proof}
In each subtask \( t_i \), which contains \( t \) operators, there are \( 2t + 1 \) tokens (as the number of numerical tokens is one more than the number of operators). Therefore, the accuracy of each subtask is given by
\begin{equation}
\label{eq:task_acc}
    P(t_i= t_i^* | H_{i-1}, q, \theta) = \left(1 - \sigma(T)\right)^{2t + 1}. 
\end{equation}

In our theoretical analysis, for simplicity, we allow \( t \) to be a fraction, defined as \( t = \frac{T}{N} \), and assume that each subtask has the same level of difficulty given \( T \) and \( N \). Under this assumption, we have the final accuracy:
     \begin{align}
        A(N) &= P(a_N=a_N^*|q,\theta) \\
        &= \prod^N_{i=1}P(t_i=t_i^*|H_{i-1},q,\theta)P(a_i=a_i^*|t_i,H_{i-1},q,\theta)\\
        &= \prod^N_{i=1}\left(1 - \sigma(T)\right)^{2t + 1}\left(1-E(N,M,T)\right)\\
        &= \left(1 - \sigma(T)\right)^{N(2t + 1)}\left(1-E(N,M,T)\right)^N\\
        &= \left(1 - \sigma(T)\right)^{2T}\left((1-E(N,M,T))(1-\sigma(T))\right)^N\\
        &= \alpha\left((1-E(N,M,T))(1-\sigma(T))\right)^N
    \end{align}
\end{proof}
\subsection{Proof of Theorem~\ref{thm:simple_optimal_N_theory}}
\lineartheory*
\begin{proof}
Given Eq.~(\ref{eq:total-acc}) that
\begin{align}
    A(N) =  \alpha\left(\left(1-\frac{T}{C}\right)\left(1-\frac{T}{NM}\right)\right)^N
\end{align}

We consider function

\begin{equation}
    f(x) \;=\; \Bigl[\bigl(1 - \tfrac{T}{Mx}\bigr)\,\bigl(1 - \tfrac{T}{C}\bigr)\Bigr]^{x}.
    \label{eq:f}
\end{equation}

For convenience, define

\[
g(x) \;=\; \ln\bigl(f(x)\bigr) \;=\; x \;\ln\!\Bigl[\bigl(1 - \tfrac{T}{Mx}\bigr)\,\bigl(1 - \tfrac{T}{C}\bigr)\Bigr].
\]

Thus,
\[
g'(x) 
= \Bigl[\ln\bigl(1 - \tfrac{T}{Mx}\bigr) \;+\; \frac{T}{Mx\,\bigl(1 - \tfrac{T}{Mx}\bigr)}\Bigr]
\;+\;
\ln\bigl(1 - \tfrac{T}{C}\bigr).
\]
Set \(g'(x) = 0\):

\[
\ln\Bigl[\bigl(1 - \tfrac{T}{Mx}\bigr)\,\bigl(1 - \tfrac{T}{C}\bigr)\Bigr]
\;+\;
\frac{T}{Mx\bigl(1 - \tfrac{T}{Mx}\bigr)} 
= 0.
\]

Let $A \;=\; \frac{1}{1 - \frac{T}{Mx}},$
then we have

\[
\ln\Bigl[\,\bigl(1 - \tfrac{T}{C}\bigr)\Bigr]
\;+\;
A-1
= \ln(A).
\]

Let \(z := 1 - T/C\). (Since \(T/C < 1\), \(z = 1 - T/C > 0\).) By moving terms, we have:

\[
-\frac{z}{e}\;=\; -A\exp(-A).
\]

Therefore, 
\[
A = -W^{-1}(-\frac{z}{e}) = -Z,
\]
Finally, we have

$$N(M,T) = x = \frac{TZ}{M(Z+1)}$$

Here \(W(\cdot)\) is the \textbf{Lambert W function}, and for \(0 < 1 - \tfrac{T}{C} < 1\), the argument \(\alpha = -\frac{1 - T/C}{e}\) lies in the interval \(\bigl(-\tfrac{1}{e}, 0\bigr)\). This means there are two real branches \(W_0\) and \(W_{-1}\) in that domain, but since $\frac{Z}{Z+1}>0$,we have $Z<-1$. Therefore, we only take the solution on branch \(W_{-1}\).
\label{pr:4.2}
\end{proof}

\subsection{Proof of Corollary~\ref{coro:scaling_law_for_linear}}
\Scalinglawsforlinear*

\begin{proof}
The second and third conclusions can be easily derived through monotonic composition, so we primarily focus on proving the first point. We begin the proof by incorporating the notation from \ref{pr:4.2}.
    We have \[
g'(x) 
= \Bigl[\ln\bigl(1 - \tfrac{T}{Mx}\bigr) \;+\; \frac{T}{Mx\,\bigl(1 - \tfrac{T}{Mx}\bigr)}\Bigr]
\;+\;
\ln\bigl(1 - \tfrac{T}{C}\bigr),
\] and $x^*(T)$ such that $g'(x^*(T))=0$.

Let $F(x^*(T),T) = g'(x^*(T)) = 0$
We want to see how \(x^*(T)\) changes as \(T\) changes, therefore we take total derivative w.r.t.\ \(T\).  By the chain rule,
\[
0 
\;=\; 
\frac{d}{dT}\,F\bigl(x^*(T),\,T\bigr)
\;=\;
\underbrace{\frac{\partial F}{\partial x}\bigl(x^*(T),T\bigr)}_{\text{call this }F_x}\;\cdot\;\frac{\partial x^*}{\partial T}\bigl(T\bigl)
\;\;+\;\;
\underbrace{\frac{\partial F}{\partial T}\bigl(x^*(T),T\bigr)}_{\text{call this }F_T}.
\] 
Hence
\[
\frac{\partial x^*}{\partial T}\bigl(T\bigl)
\;=\;
-\;\frac{F_T\Bigl(x^*(T),\,T\Bigr)}{\,F_x\,\Bigl(x^*(T),\,T\Bigr)}.
\]
So the sign of \(x'^*(T)\) is the opposite of the sign of \(F_T\), provided \(F_x\neq 0\).  

Since
\begin{equation}
    \,F_x\,\Bigl(x,\,T\Bigr) = -\frac{T^2}{x(Mx-T)^2} < 0, \forall x > 0,
\end{equation}

all we need to prove is 

\begin{equation}
    \,F_T\,\Bigl(x^*(T),\,T\Bigr) = \frac{T}{(Mx^*(T)-T)^2}-\frac{1}{C-T}>0.
\end{equation}

That is 

\begin{equation}
    \frac{\sqrt{T(C-T)}+T}{M}>x^*(T).
\end{equation}

Let $x_0(T) = \frac{\sqrt{T(C-T)}+T}{M}$ be the test point.

According to Lemma~\ref{lm:test_point}, $F(x_0(T),T)<0$. Since $F(x^*(T),T) = 0, \text{ and } \,F_x\,\Bigl(x^*(T),\,T\Bigr)<0, $ we have $x_0(T)>x^*(T)$.

Thus, $\,F_T\,\Bigl(x^*(T),\,T\Bigr)>0$ holds and we have proved our corollary with $\frac{\partial x^*}{\partial T}\bigl(T\bigl) > 0$.

\end{proof}

\subsection{Proof of Theorem~\ref{thm:general_theory}}
\general*

\begin{proof}
    (1) Since $0<A(N)<(1-\sigma(T))^N$, and $\lim_{N\rightarrow +\infty} (1-\sigma(T))^N = 0$, $\lim_{N\rightarrow +\infty} A(N,M,T) = 0$

    (2) Let $g(x)$ denote $E(x,M,T)$ and define $f(x) = \ln A(x)$. Then,  
\begin{align}
    f'(x) 
    &= \ln (1-\sigma(T)(1-g(x)))  - \frac{xE'(x)}{1-E(x)}\\
    &< \ln (1-\sigma(T)(1-g(x)))  + 2, \quad \text{(since $E$ is convex and $x = N \geq 1$)}
\end{align}
If $A(N)$ attains its maximum at some point $N^*>1$, then $\ln (1-\sigma(T))  + 2 > 0$. Otherwise, we would have $f'(x)<\ln (1-\sigma(T))  + 2 \leq 0 \;\forall x>1$, leading to a contradiction. 

Thus, it follows that $e^2(1-\sigma(T))>1$. 

Now, define $N(M,T) = E^{-1}\left(1-\frac{1}{e^2(1-\sigma(T))}\right)$, which satisfies  
\[
\ln (1-\sigma(T)(1-g(N(M,T))))  + 2 = 0.
\]
If there exists $x^*<N(M,T)$ such that $f'(x^*) = 0$, then we obtain  
\[
0 = f'(x^*)< \ln (1-\sigma(T)(1-E(x)))  + 2<0,
\]
which is a contradiction. Hence, the assumption that $x^*<N(M,T)$ must be false. 

Therefore, we conclude that $x^* = N^* > N(M,T)$.

\end{proof}

\subsection{Proof of Theorem \ref{thm:random_error}}
\random*
\begin{proof}
According to the multidimensional version of Hölder's inequality, 
\begin{align}
    \mathbb{E}\left[\prod_{i=1}^N(1-e_i)(1-\sigma_i)\right] &\leq \prod_{i=1}^N \left( \mathbb{E}[(1 - e_i)^{2N}] \mathbb{E}[(1 - \sigma_i)^{2N}]\right)^{\frac{1}{2N}}\\
    \text{(Lemma \ref{lm:moment})}&\leq \prod_{i=1}^N \left(1- \frac{T}{C+2N-1}\right) \left(1- \frac{T}{NM+2N-1}\right)\\
    &= \left[\left(1- \frac{T}{C+2N-1}\right) \left(1- \frac{T}{NM+2N-1}\right)\right]^{N}
\end{align}

\end{proof}
\subsection{Proof of Corollary~\ref{cor:rl_converge}}

\rlconverge*
\begin{proof}
We treat the choice of CoT length as a $k$-armed stochastic bandit with
action set $\mathcal A=\{N_1,\dots,N_k\}$ and unknown success
probabilities\footnote{By Proposition~\ref{pro:general_acc}, $A(N_i)$ is the
probability that the final answer is correct when a chain of length $N_i$
is used.  The bandit is \emph{stationary} because $A(N_i)$ does not depend
on time or the agent’s past actions.}
$A(N_i)\in(0,1)$.
Without loss of generality, relabel the arms so that
\[
A(N_1)=\max_{j}A(N_j)\eqqcolon A^*,\qquad
A(N_1)\ge A(N_2)\ge\cdots\ge A(N_k).
\]

The agent uses a softmax (Gibbs) policy
\begin{equation}
    \pi_\theta(N_i)\;=\;\frac{e^{\theta_i}}{\sum_{j=1}^{k}e^{\theta_j}},
    \qquad \theta\in\mathbb R^{k},
    \label{eq:softmax_policy}
\end{equation}
and maximises the expected reward
\begin{equation}
    J(\theta)\;=\;\sum_{i=1}^{k}\pi_\theta(N_i)\,A(N_i).
    \label{eq:objective}
\end{equation}
Because $\pi_\theta$ is $C^{\infty}$ in $\theta$ and $A(N_i)$ are constants,
$J$ is smooth.

Under the REINFORCE estimator with sufficiently small, fixed step size
$\eta>0$,
gradient ascent updates take the form
\begin{equation}
    \theta^{(t+1)}\;=\;\theta^{(t)}
    +\eta\,\nabla_\theta J\bigl(\theta^{(t)}\bigr),
\end{equation}
where$\vphantom{\bigl|}$
\begin{equation}
    \frac{\partial J}{\partial\theta_i}
    \;=\;\pi_\theta(N_i)\Bigl(A(N_i)-J(\theta)\Bigr).
    \label{eq:grad_comp}
\end{equation}
Eq.~(\ref{eq:grad_comp}) is the classical \emph{replicator} (or logit) gradient.
Define the simplex
$\Delta^{k-1}\!:=\!\{\pi\in(0,1]^{k}\mid\sum_i\pi_i=1\}$ and write
$\pi_\theta=(\pi_\theta(N_1),\dots,\pi_\theta(N_k))$.

Letting $\eta\to0$ yields the ODE
\begin{equation}
    \dot{\pi}_i
    \;=\;\pi_i\bigl(A(N_i)-\langle \pi,A\rangle\bigr),
    \qquad i=1,\dots,k,
    \label{eq:replicator}
\end{equation}
with $\langle \pi,A\rangle=\sum_j\pi_jA(N_j)$.
Eq.~(\ref{eq:replicator}) is the \textbf{replicator dynamics}
for a fitness landscape $A$ on $\Delta^{k-1}$.

Consider the Kullback–Leibler divergence to the optimal pure strategy
$\mathbf e_1=(1,0,\dots,0)$,
\[
    V(\pi)\;=\;\sum_{i=1}^{k}\pi_i\ln\!\Bigl(\tfrac{\pi_i}{e_{1,i}}\Bigr)
    \;=\;-\ln \pi_1.
\]
$V$ is non-negative on $\Delta^{k-1}$ and $V(\pi)=0$ iff $\pi=\mathbf e_1$.

Taking the time derivative along Eq.~(\ref{eq:replicator}) gives
\[
\frac{dV}{dt}
= -\frac{\dot{\pi}_1}{\pi_1}
= -\bigl(A(N_1)-\langle\pi,A\rangle\bigr)
\;\le\;0,
\]
with equality iff $\pi_1=1$ \emph{or} $A(N_1)=\langle\pi,A\rangle$.
The latter can only happen if $\pi_1=1$ because $A(N_1)>A(N_j)$ for $j>1$.
Hence $V$ is a strict Lyapunov function, and $\mathbf e_1$ is the
\emph{unique} asymptotically stable equilibrium of Eq.~(\ref{eq:replicator}).
All other stationary points (mixtures over sub-optimal arms) are unstable.

For sufficiently small but fixed $\eta$
(choose $\eta<\frac{1}{A^*}$, which always exists),
projected gradient ascent is a
\emph{perturbed} discretisation of Eq.~(\ref{eq:replicator}).
Standard results for primal-space mirror descent imply that the discrete iterates
$\pi^{(t)}\equiv\pi_{\theta^{(t)}}$
converge almost surely to the set of asymptotically stable equilibria of the
ODE, i.e.\ to $\{\mathbf e_1\}$.
Therefore
\[
\lim_{t\to\infty}\pi_{\theta^{(t)}}(N_i)
=\begin{cases}
1, &\text{if } i=\arg\max_j A(N_j),\\[4pt]
0, &\text{otherwise.}
\end{cases}
\]
Because $A$ may attain its maximum at several arms, the limit is a deterministic
policy that places all probability on \emph{some} maximiser of $A$.

Thus gradient ascent on Eq.~(\ref{eq:objective}) converges to a deterministic
policy that always selects an optimal CoT length
$N^*=\arg\max_{N\in\mathcal A}A(N)$, completing the proof.
\end{proof}

\subsection{Technical Lemmas}
\begin{lemma}[test point]
\label{lm:test_point}
    Let \( F(x) \) be defined as  
    \[
    F(x) = \ln\left(1 - \frac{T}{Mx}\right) + \frac{T}{Mx\left(1 - \frac{T}{Mx}\right)} + \ln\left(1 - \frac{T}{C}\right),
    \]
    where \( T, M, C \in \mathbb{R}^+ \) satisfy the conditions:
    \begin{itemize}
        \item \( 0 < \frac{T}{C} < 0.9 \),
        \item \( 0 < \frac{T}{Mx} < 1 \).
    \end{itemize}
    
    Define \( x_0 \) as
    \[
    x_0 = \frac{\sqrt{T(C-T)}+T}{M}.
    \]
    Then, we have
    \[
    F(x_0) < 0.
    \]
\end{lemma}

\begin{proof}
    At \( x = x_0 \), note that
\[
Mx_0 = \sqrt{T(C-T)} + T.
\]

Thus,
\[
1 - \frac{T}{Mx_0} 
= 1 - \frac{T}{T+\sqrt{T(C-T)}}
= \frac{\sqrt{T(C-T)}}{T+\sqrt{T(C-T)}}.
\]

Therefore,
\[
\ln\Bigl(1-\frac{T}{Mx_0}\Bigr)
=\ln\left(\frac{\sqrt{T(C-T)}}{T+\sqrt{T(C-T)}}\right)
=\ln\sqrt{T(C-T)} - \ln\bigl(T+\sqrt{T(C-T)}\bigr).
\]

Also, observe that
\[
\frac{T}{Mx_0\Bigl(1-\frac{T}{Mx_0}\Bigr)}
=\frac{T}{(T+\sqrt{T(C-T)})\Bigl(\frac{\sqrt{T(C-T)}}{T+\sqrt{T(C-T)}}\Bigr)}
=\frac{T}{\sqrt{T(C-T)}}
=\sqrt{\frac{T}{C-T}}.
\]

It is convenient to introduce the change of variable
\[
u = \sqrt{\frac{T}{C-T}},
\]
so that
\[
T = u^2 (C-T),\quad \sqrt{T(C-T)} = u(C-T). 
\]
Then we have
$$T+\sqrt{T(C-T)} = u^2 (C-T) + u(C-T) = u(C-T)(u+1).$$

In these terms we have:
\[
\ln\sqrt{T(C-T)} = \ln\bigl[u(C-T)\bigr] = \ln u + \ln(C-T),
\]
\[
\ln\bigl(T+\sqrt{T(C-T)}\bigr) = \ln\bigl[u(C-T)(u+1)\bigr] = \ln u + \ln(C-T) + \ln(u+1),
\]
and
\[
\sqrt{\frac{T}{C-T}} = u.
\]

Finally, we have
\[
\ln\Bigl(1-\frac{T}{C}\Bigr) = -\ln(\frac{C}{C-T}) = -\ln(u^2 +1)
\]

Thus, the function \( F(x_0) \) becomes
\begin{align}
F(x_0)
&=\ln u + \ln(C-T) - \bigl(\ln u + \ln(C-T) + \ln(u+1)\bigr) + u -\ln (u^2+1)\\
&= -\ln(u+1) + u - \ln(u^2 +1)\, ,
\end{align}
where $u = \sqrt{\frac{T}{C-T}} \in \left(0,3\right).$
It is easy to show $F(x_0)<0$ when $u \in \left(0,3\right)$.
\end{proof}

\begin{lemma}[Estimation of the $n$-th Moment of the Beta Distribution]
\label{lm:moment}
Let $x \sim \mathrm{Beta}(\alpha, \beta)$. Then
\[
\mathbb{E}[(1 - x)^n] \leq \left(1 - \frac{\alpha}{\alpha + \beta + n - 1}\right)^n.
\]
\end{lemma}

\begin{proof}
\begin{align*}
    \mathbb{E}[(1 - x)^n] 
    &= \frac{1}{B(\alpha, \beta)} \int_0^1 (1 - x)^n x^{\alpha - 1} (1 - x)^{\beta - 1} \, dx \\
    &= \frac{1}{B(\alpha, \beta)} \int_0^1 x^{\alpha - 1} (1 - x)^{\beta + n - 1} \, dx \\
    &= \frac{B(\alpha, \beta + n)}{B(\alpha, \beta)} \\
    &= \frac{\Gamma(\alpha) \Gamma(\beta + n)}{\Gamma(\alpha + \beta + n)} \cdot \frac{\Gamma(\alpha + \beta)}{\Gamma(\alpha) \Gamma(\beta)} \\
    &= \frac{\Gamma(\beta + n)}{\Gamma(\beta)} \cdot \frac{\Gamma(\alpha + \beta)}{\Gamma(\alpha + \beta + n)} \\
    &= \prod_{i = 0}^{n - 1} \frac{\beta + i}{\alpha + \beta + i} \\
    &\leq \left( \frac{\beta + n - 1}{\alpha + \beta + n - 1} \right)^n \\
    &= \left(1 - \frac{\alpha}{\alpha + \beta + n - 1} \right)^n.
\end{align*}
\end{proof}

\section{Pseudo-code of Length-filtered Vote}

\label{app:length_filtered_vote}

\begin{algorithm}[H]
\caption{Length-filtered Vote}
\label{alg:la-vote-top-k}
\begin{algorithmic}[1]
\State \textbf{Input:} Model $f_\theta$, Question $q$, Space of All Possible Answers $A$, Number of Total Groups $M$, Number of Selected Groups $K$, Group Width $D$
\State \textbf{Output:} Final Answer $\hat{a}$

\State Sample candidates $c_1, \dots, c_n \stackrel{i.i.d.}{\sim} f_\theta(q)$
\State \textbf{Define} $\mathcal{A}(c)$ as the corresponding answer of candidates $c$.
\State \textbf{Define} $p_j \in [0,1]^{|\mathcal{A}|}$ as the frequency of each answer in length group $L_j$.
\For{$j = 1$ to $m$}

    $L_j = \{c_i \mid \ell(c_i) \in \left[D*(j-1),D*j\right) , i = 1,\cdots,n \}$

    \For{$a \in \mathcal{A}$}
        \[
        p_j[a] = \frac{\sum_{c \in L_j} \mathbb{I}(\mathcal{A}(c) = a)}{|L_j|}
        \]
    \EndFor
\EndFor

\State$
\{s_1, \dots, s_K\} = \arg\min_{S \subseteq \{1, \dots, M\}, |S|=K} \sum_{s \in S} H(p_s)
$

\State$
\hat{a} = \arg\max_{a \in A} \sum_{c \in L_{s_1} \cup \dots \cup L_{s_K}} \mathbb{I}(\mathcal{A}(c)=a)
$

\State \Return $\hat{a}$
\end{algorithmic}
\end{algorithm}

\end{document}